\documentclass[a4paper,fleqn]{cas-sc}

\usepackage{graphicx}
\usepackage[tight,footnotesize]{subfigure}
\usepackage{color}
\usepackage{array}
\usepackage{tabularx}
\usepackage[strict]{changepage}
\usepackage{amssymb}
\usepackage{pifont}
\usepackage{enumitem}
\usepackage{gensymb}
\usepackage{makecell}
\usepackage{textcomp}
\usepackage{float}
\usepackage{balance}
\usepackage{mathtools}
\usepackage{ctable} 
\usepackage{amsthm}

\usepackage{algorithm}
\usepackage{algorithmic}
\usepackage{arydshln}

\usepackage[square,numbers]{natbib}

\newcolumntype{L}[1]{>{\raggedright\let\newline\\\arraybackslash\hspace{0pt}}m{#1}}
\newcolumntype{C}[1]{>{\centering\let\newline\\\arraybackslash\hspace{0pt}}m{#1}}
\newcolumntype{R}[1]{>{\raggedleft\let\newline\\\arraybackslash\hspace{0pt}}m{#1}}

\def\tsc#1{\csdef{#1}{\textsc{\lowercase{#1}}\xspace}}
\tsc{WGM}
\tsc{QE}

\begin{document}
\let\WriteBookmarks\relax
\def\floatpagepagefraction{1}
\def\textpagefraction{.001}

\shorttitle{ProNet for Non-AutoRegressive Multi-Horizon Time Series Forecasting}    


\title [mode = title]{Progressive Neural Network for Multi-Horizon Time Series Forecasting}

\tnotemark[1] 


%
%
\author[1]{Yang Lin}[orcid=0000-0003-4168-6750]

\cormark[1]

\fnmark[1]

\ead{ylin4015@uni.sydney.edu.au}

\ead[url]{https://scholar.google.com/citations?user=T00NBYIAAAAJ&hl=en}
%
%
\affiliation[1]{organization={The University of Sydney},
            city={Sydney},
            state={NSW},
            country={Australia}}
%
%
%
%
%
%
%
%



\begin{abstract}
 In this paper, we introduce ProNet, an novel deep learning approach designed for multi-horizon time series forecasting, adaptively blending autoregressive (AR) and non-autoregressive (NAR) strategies. Our method involves dividing the forecasting horizon into segments, predicting the most crucial steps in each segment non-autoregressively, and the remaining steps autoregressively. The segmentation process relies on latent variables, which effectively capture the significance of individual time steps through variational inference.
		In comparison to AR models, ProNet showcases remarkable advantages, requiring fewer AR iterations, resulting in faster prediction speed, and mitigating error accumulation. On the other hand, when compared to NAR models, ProNet takes into account the interdependency of predictions in the output space, leading to improved forecasting accuracy.
		Our comprehensive evaluation, encompassing four large datasets, and an ablation study, demonstrate the effectiveness of ProNet, highlighting its superior performance in terms of accuracy and prediction speed, outperforming state-of-the-art AR and NAR forecasting models.
 
\end{abstract}


	 


\maketitle


\section{Introduction}

Time series forecasting has a wide range of applications in industrial domain for decades, including predicting electricity load, renewable energy generation, stock prices and traffic flow, air quality \cite{WANG2022262} etc. Many methods have been developed for this task that can be classified into two broad categories. 
In the early years, statistical models such as Auto-regressive Integrated Moving Average (ARIMA) and State Space Models (SSM) \cite{Rob2019book} were widely used by industry forecasters. However, they fit each time series independently and are not able to infer shared patterns from related time series \cite{Logsparse19NIPS}. 
{On the other hand, machine learning methods have been developed for modelling the non-linearity from time series data. Preliminary methods are random forest \cite{QIU2017249}, Support Vector Machine (SVM) \cite{CHEN201599} and Bayesian methods \cite{DU2022155}. 
Moreover, recent research has widely acknowledged the effectiveness of time series decomposition and ensemble learning methods in refining forecasting models \cite{LV2022994,GAO2022117784,DU2022155,WANG2023122504}. 
Ensemble learning methods have gained recognition for their ability to combine individual models and enhance overall predictive performance while minimizing overfitting. Du et al. \cite{DU2022155} develop the ensemble strategy that takes advantage of high diversification statistical, machine learning and deep learning methods, and assigns time-varying weights for model candidates with bayesian optimization, to avoid the shortage of model choice and alleviates the risk of underfitting. Similarly, Gao et al. \cite{GAO202351}  introduced an online dynamic ensemble of deep random vector functional link with three stages for improved performance. 
Decomposition-based methods have also shown promise in time series forecasting by breaking down the data into underlying components, leading to more accurate and manageable predictions. Different decomposition approaches, such as classical decomposition, moving averages, and state space model have been explored. 
For instance,  Li et al. \cite{LI2023833} proposed a convolutional neural network ensemble method that leverages decomposed time series and batch normalization layers to reduce subject variability.  
Wang et al. \cite{9763011} proposed a fuzzy cognitive map to produce interpretable results by forecasting the decompositional components: trend, fluctuation range, and trend persistence.  
Lin et al. \cite{9679135} developed SSDNet, employing Transformer architecture to estimate state space model parameters and provide time series decomposition components: trend and seasonality. Tong et al. \cite{doi:10.1137/1.9781611977653.ch54,TONG2023119410} ntroduced Probabilistic Decomposition Transformer with hierarchical mechanisms to mitigate cumulative errors and a conditional generative approach for time series decomposition. Furthermore, Wang et al. \cite{WANG2023122504} introduced the ternary interval decomposition ensemble learning method, addressing limitations of point and interval forecasting models.
The amalgamation of machine learning models, time series decomposition, and ensemble learning has demonstrated great promise as a potent solution for advancing forecasting performance. Notably, the philosophy of decomposition and ensemble can seamlessly integrate with major machine learning models, further enhancing their effectiveness in various applications. 
} 

Recently, a sub-class of machine learning methods - deep learning, has been widely studied for forecasting tasks due to their strong ability of modelling complex and related dependencies from large-scaled time series. 
Existing deep learning methods can be divided into  AutoRegressive (AR) and Non-AutoRegressive (NAR) models on the perspective of how they make the multi-step forecasts.  
{Notable examples of AR models include DeepAR \cite{DeepAR20}, DeepSSM \cite{DeepSSM18NIPS}, DeepFactor \cite{DeepFactors}, DAttAE \cite{PHAM2022117514}, LogSparse Transformer \cite{Logsparse19NIPS}, visibility graph model \cite{GAO2022553}, RDNM-ANN \cite{EGRIOGLU2022572} and TimeGrad \cite{rasul2021autoregressive}.  NAR - MQ-RNN \cite{MQRNN18}, N-BEATS \cite{N-BEATS}, AST \cite{AST20NIPS} and Informer \cite{Informer21} are the prominent NAR methods. }

AR forecasting models have the problem of slow inference speed and error accumulation due to the use of a recursive method that use previously predicted values to make future forecasts. AR models are usually trained with the teacher-forcing mechanism and consider ground truth as previous predictions to feed into the model during training. This causes a discrepancy between training and prediction, and could cause unsatisfied accuracy for long forecasting horizons 
\cite{bengio2015scheduled, AST20NIPS}. 
In contrast, NAR forecasting models overcome the aforementioned problems since they generate all predictions within forecasting horizon simultaneously. 
However, NAR model ignores interdependencies in output space and such assumption violates real data distribution for sequence generation tasks \cite{Gu18ICLR,Gu20Tricks}. This may result in unrelated forecasts over the prediction horizon and accuracy degradation \cite{Taieb16,AST20NIPS}. 
Empirically, AR methods were found to be better for shorter horizons but outperformed by NAR for longer horizons due to error accumulation \cite{Taieb16}. 
Thus, both AR and NAR models have their own complementary strengths and limitations for multi-horizon forecasting which stem from their prediction strategy. 
Recently NAR models have been proposed specific for translation tasks that can alleviate the accuracy degradation by performing dependency reduction in output space and reduce the difficulty of training \cite{Gu18ICLR,barezi2020study,Ren20ACL,Gu20Tricks}. However, such studies are scarce for time series forecasting tasks. 

\begin{figure}[t]
	\centering	  
	\includegraphics[width=.6\columnwidth]{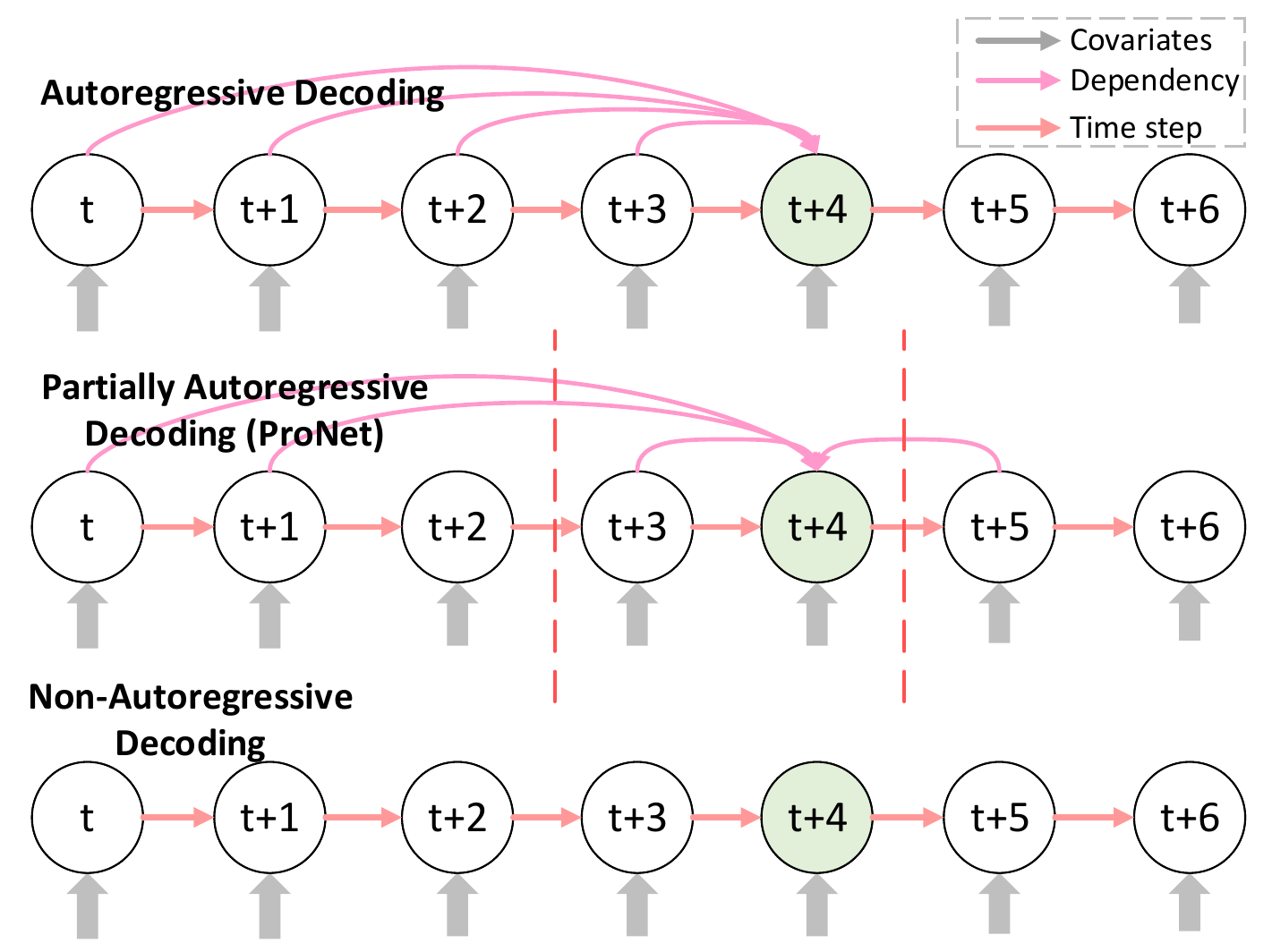}
	\caption{Illustration of AR, ProNet partially AR and NAR decoding process: 1) AR decoder forecasts with covariates and all previous predictions; 2) NAR decoder forecasts all steps with covariates only in parallel; 3) our partially AR decoder divides horizon into segments (indicated by red dash lines), individual each segment is predicted autoregressively with covariates and previous predictions of all segments, while each prediction of segments can be made simultaneously. }
	\label{Connects}
\end{figure}

{
	
A balance must be struck between AR and NAR forecasting models to tackle the challenges of error accumulation and low latency in AR models, alongside the NAR models' inability to adequately capture interdependencies within the output space.
Recent strides in this domain have illuminated the advantages of incorporating dependency and positional information within the prediction horizon. These breakthroughs have exhibited their efficacy across a spectrum of sequence modeling tasks. For instance, Ran et al. \cite{SemiAR20ACL} have ingeniously integrated future predictions to overcome the multi-modality predicament in neural machine translation. In a parallel vein, Fei \cite{Fei21AAAI} and Zhou et al. \cite{Zhou21Semi} have skillfully amalgamated information from future time steps to generate past predictions, exemplified in the context of caption generation.
Furthermore, Han et al. \cite{han-etal-2023-ssd} have introduced a diffusion-based language model with bidirectional context updates, adding a notable dimension to the evolving landscape of research in this field. 
To address these challenges and capitalize on the strengths of both AR and NAR modeling, we introduce Progressive Neural Network (ProNet), a novel deep learning approach designed for time series forecasting. ProNet strategically navigates the AR-NAR trade-off, leveraging their respective strengths to mitigate error accumulation and slow prediction while effectively modeling dependencies within the target sequence. 
Specifically, ProNet adopts a partially AR prediction strategy by segmenting the forecasting horizon. It predicts a subset of steps within each segment using a non-autoregressive approach, while maintaining an autoregressive decoding process for the remaining steps. 

Fig. \ref{Connects} illustrates the different prediction mechanism of AR, ProNet's partially AR, and NAR decoding mechanisms. For example, when the AR decoder considers step $t+4$ dependent on steps from $t$ to $t+3$, the NAR decoder assumes no dependency. In contrast, ProNet's partially AR decoder takes into account dependencies from past steps $t$, $t+1$, $t+3$, as well as future step $t+5$.
The initiation of horizon segments is determined by latent variables, optimizing their training through variational inference to capture the significance of each step. Consequently, in comparison to AR models, ProNet's predictions require fewer iterations, enabling it to overcome error accumulation while achieving faster testing speeds. Moreover, compared to NAR models, ProNet excels in capturing dependencies within the target space.
}

The main contributions of our work are as follows:
\begin{enumerate}
	\item We propose ProNet, a partially AR time series forecasting approach that generates predictions of multiple steps in parallel to leverage the strength of AR and NAR models. 
	Our ProNet assumes an alternative dependency in target space and incorporates information of further future to generate forecasts. 
	
	\item We evaluate the performance of ProNet on four time series forecasting tasks and show the advantages of our model against most state-of-the-art AR and NAR methods with fast and accurate forecasts. An ablation study confirmed the effectiveness of the proposed horizon dividing strategy.
\end{enumerate}

\section{Related Work}


Recent advancements in forecasting methodologies have led to the emergence of NAR forecasting models \cite{MQRNN18,N-BEATS,AST20NIPS,Informer21}. These models seek to address the limitations of AR models by eschewing the use of previously generated predictions and instead making all forecasts in a single step. However, the effectiveness of NAR forecasting models is hindered by their assumption of non-interdependency within the target space. This assumption arises from the removal of AR connections from the decoder side, leading to the estimation of separate conditional distributions for each prediction independently \cite{Gu18ICLR,barezi2020study,Ren20ACL,Gu20Tricks}. While both AR and NAR models have proven successful in forecasting applications, AR methods tend to excel for shorter horizons, while NAR methods outperform AR for longer horizons due to error accumulation \cite{Taieb16}. 
Unlike AR models, NAR models offer the advantage of parallelizable training and inference processes. However, their output may present challenges due to the potential generation of unrelated forecasts across the forecast horizon. This phenomenon could lead to discontinuous and unrealistic forecasts \cite{Taieb16}, as the incorrect assumption of independence prevents NAR models from effectively capturing interdependencies between each prediction.

Serval research \cite{Gu18ICLR,barezi2020study,Ren20ACL,Gu20Tricks} have been made to enhance NAR models, although most of these efforts have been focused on Neural Machine Translation (NMT) tasks. Gu et al. \cite{Gu18ICLR} introduced the NAR Transformer model, which reduces output dependencies by incorporating fertilities and leveraging sequence-level knowledge distillation techniques \cite{KD15,Kim16EMNLP}. Recent developments have seen the adaptation of NAR models for translation tasks, mitigating accuracy degradation by tackling output space dependencies. This approach aims to capture and manage dependencies, thereby alleviating training challenges \cite{Gu18ICLR,barezi2020study,Ren20ACL}. Notably, knowledge distillation \cite{KD15,Kim16EMNLP} emerges as a highly effective technique to enhance NAR model performance.

\begin{figure}[!t]
	\centering	  
	\subfigure[]{\includegraphics[width=.3\columnwidth]{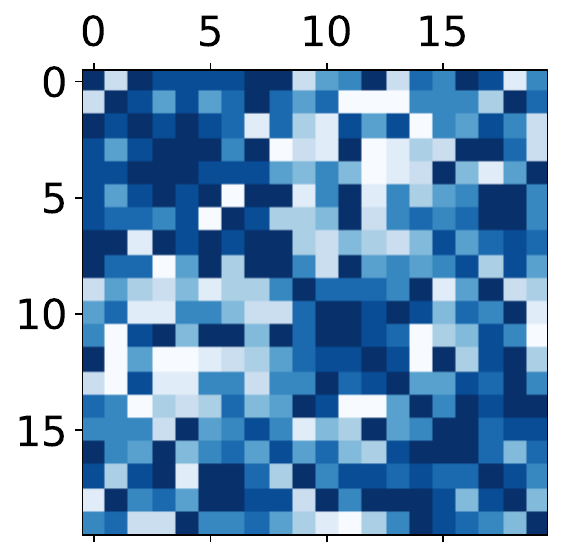}}		\subfigure[]{\includegraphics[width=.3\columnwidth]{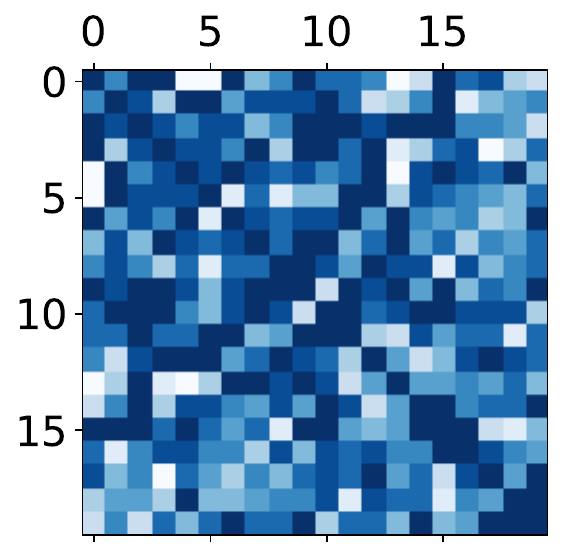}}
	\caption{Partial correlation of Sanyo set for two different days (20 time steps for each day).}
\label{elect_pc}
\end{figure}

The trade-off between AR and NAR \cite{SemiAR18EMNLP,SemiAR20ACL,Fei21AAAI,Zhou21Semi}  has been a subject of exploration, particularly in the context of NMT and other sentence generation tasks. Notable instances include the works of \cite{SemiAR18EMNLP,Zhou21Semi}, which retain AR properties while enabling parallel prediction of multiple successive words. Similarly, \cite{SemiAR20ACL,Fei21AAAI} employ a strategy that generates translation segments concurrently, each being generated autoregressively. 
However, prior approaches have relied on dividing the target sequence into evenly distributed segments, assuming fixed dependencies among time steps. This assumption, while applicable in some contexts, proves unsuitable for time series forecasting due to the dynamic and evolving nature of real-world time series data. For instance, in Fig. \ref{elect_pc}, we visualize the partial correlation of two distinct days (comprising 20 steps each) from the Sanyo dataset. Evidently, the two plots exhibit varying dependency patterns, signifying that the most influential time steps differ between the two cases. 
Additionally, it becomes apparent that future steps can exert substantial influence on preceding steps. Take Fig. \ref{elect_pc} (a) as an example, where step 5 exhibits strong partial correlation with steps 17 and 18. This correlation suggests that incorporating information from steps 17 and 18 while predicting step 5 could be highly beneficial.

In this work, we present ProNet that navigates the intricate balance between AR and NAR models. 
We extend previous work with several key enhancements: 1) assuming a non-fixed dependency pattern and identifying the time steps that need to be predicted first via latent factors and then predict further groups of steps autoregressively; 
2) assuming the alternative time-varying dependency and incorporating future information into forecasting; 
3) introducing the sophisticated-designed masking mechanism to train the model non-autoregressively.

\section{Problem Formulation}


Given is: 1) a set of $N$ univariate time series (solar or electricity series) $\{\mathbf{Y}_{i,1:T_l}\}^N_{i=1}$, where $\mathbf{Y}_{i,1:T_l}\coloneqq[{y}_{i,1}, {y}_{i,2},...,{y}_{i,T_l}]$, $T_l$ is input sequence length, and ${y}_{i,t}\in\Re$ is value of the $i$th time series at time $t$; 2) a set of associated time-based multi-dimensional covariate vectors $\{\mathbf{X}_{i, 1: T_l+T_h}\}_{i=1}^{N}$, where $T_h$ is  forecasting horizon length and $T_l+T_h=T$. 
Our goal is to predict the future values of the time series $\{\mathbf{Y}_{i,T_l+1:T_l+T_h}\}^N_{i=1}$, i.e. the PV power or electricity usage for the next $T_h$ time steps after $T_l$. 

AR forecasting models produce the conditional probability of the future values:
\begin{equation}
\begin{split}
	&p\left(\mathbf{Y}_{i,T_l+1:T_l+T_h} \mid \mathbf{Y}_{i,1:T_l}, \mathbf{X}_{i, 1: T_l+T_h} ; \theta\right)
	\\ =& \prod_{t=T_{l}+1}^{T_{l}+T_{h}} p\left({y}_{i, t} \mid \mathbf{Y}_{i, 1: t-1}, \mathbf{X}_{i, 1: t} ; \theta\right),
\end{split}
\end{equation}
where the input of model at step $t$ is the concatenation of ${y}_{i, t-1}$ and ${x}_{i, t}$ and $\theta$ denotes the model parameters. 

For NAR forecasting models, the conditional probability can be modelled as:
\begin{equation}
\begin{split}
	&p\left(\mathbf{Y}_{i,T_l+1:T} \mid \mathbf{Y}_{i, 1: T_l}, \mathbf{X}_{i, 1: T} ; \theta\right) 
	\\=& \prod_{t=T_{l}+1}^{T} p\left({y}_{i, t} \mid \mathbf{Y}_{i, 1: T_l}, \mathbf{X}_{i, 1: T} ; \theta\right)
\end{split}
\end{equation}

{
Table \ref{featuresARNAR} presents a comparison of available information for predicting step $t+1$ using AR and NAR forecasting methods. Both AR and NAR methods have access to covariates and ground truth from the past. However, there is a distinction in the scope of information they can utilize. The AR method can only make use of covariates and ground truth up to time step $t$, whereas NAR methods can utilize all covariates within the forecasting horizon but do not have access to ground truth.
}
\begin{table*}[t]
	\renewcommand{\arraystretch}{1}	
	\centering
	\begin{tabular}{C{2cm}  C{3cm} C{3cm} C{3cm} C{3cm} } 
		\specialrule{.1em}{.05em}{.05em} 
		Prediction at $t$&\multicolumn{4}{c}{Input at $t$} \\	
		\hline
		&past covariates &future covariates&past ground truth &future ground truth\\	
		\hline
		AR ($\mathbf{Y}_{i, t+1}$)&$\mathbf{X}_{i, 1: T_l}$& $\mathbf{X}_{i, T_l : t}$& $\mathbf{Y}_{i, 1: T_l}$&$\mathbf{Y}_{i, T_l : t}$\\				
		NAR ($\mathbf{Y}_{i, t+1}$)&$\mathbf{X}_{i, 1: T_l}$& $\mathbf{X}_{i, T_l : T_l +T_h}$& $\mathbf{Y}_{i, 1: T_l}$& None\\			
		\specialrule{.1em}{.05em}{.05em} 
	\end{tabular}
	\caption{Available information for predicting step $t+1$ by AR and NAR forecasting methods. }
	\label{featuresARNAR} 
\end{table*}

Specifically, ProNet produces the conditional probability distribution of the future values, given the past history: $p\left(\mathbf{Y}_{i,T_l+1:T} \mid \mathbf{Y}_{i, 1: T_l}, \mathbf{X}_{i, 1: T} ; \theta\right)$, where the input of model at step $t$ is the concatenation of ${y}_{i, t-1}$ and ${x}_{i, t}$ and $\theta$ denotes the model parameters. 

The models are applicable to all time series, so the subscript $i$ will be omitted in the rest of the paper for simplicity.

\section{Progressive Neural Network}
In this section, we first present the architecture of ProNet and then explain its details in four sections: 1) \textit{partially AR forecasting mechanism} to overcome the limitations of AR and NAR decoders, 2) \textit{progressive forecasting} to correct inaccurate predictions made at early stages, 3) \textit{progressive mask} that implements the previous two mechanisms for Transformer model and 4) \textit{variational inference} to generate the latent variables with dependency information to serve the partially AR forecasting mechanism.

\subsection{Architecture}


\begin{figure}[t]
\centering	  
\includegraphics[width=.7\columnwidth]{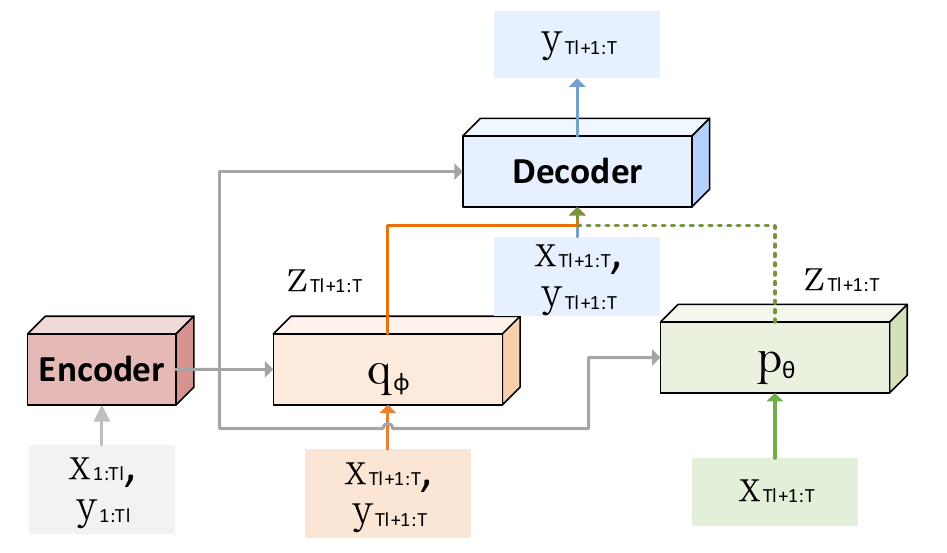}
\caption{Structure of the four components in ProNet: encoder, decoder, prior model $p_{\theta}$ and posterior model $q_{\phi}$.}
\label{ProNet}
\end{figure}

{
Figure. \ref{ProNet} illustrates the architecture of ProNet, a partially AR time series forecasting model by using latent variables to model the uncertainty in target space. 
ProNet comprises four core components: an encoder, a forecasting decoder, a prior model denoted as $p_{\theta}({z} \mid {x})$, and a posterior model denoted as $q_{\phi}({z} \mid {y}, {x})$. 

During each training iteration, the feedforward process unfolds through four stages:

\begin{enumerate}
	\item Encoder for Pattern Extraction: The encoder analyzes patterns from preceding time steps, contributing valuable insights to all decoders.
	\item Significance Assessment by Posterior Model: The posterior model $q_{\phi}({z} \mid {y}, {x})$ integrates both ground truth and covariates, effectively discerning the significance of time steps within the forecasting horizon. This assessment identifies pivotal steps, subsequently used to segment the forecasting horizon.
	\item Significance Assessment by Prior  Model: A separate prior model $p_{\theta}({z} \mid {x})$ employs covariates to predict the importance of time steps within the horizon. The outputs of this prior model are meticulously calibrated to closely approximate the posterior model's outcomes.
	\item Decoding and Forecast Generation: The decoder $p(y\mid x,z)$ employs the ground truth, covariates, and the output of the posterior model $q_{\phi}({z} \mid {y}, {x})$ to segment the forecasting horizon into distinct segments for accurate forecast generation.
\end{enumerate}

During the inference phase, the posterior model is omitted, and the prior model seamlessly takes on its role, facilitating accurate predictions. Notably, in the absence of ground truth during prediction, the decoder employs past predictions to generate forecasts.

As the architectural backbone of ProNet, we adopt the Informer architecture \cite{Informer21}; however, it is pertinent to highlight that alternative Transformer-based architectures can be seamlessly integrated into the ProNet framework. Impressively, ProNet's efficacy remains pronounced when employing a vanilla Transformer as its architectural backbone.

In summary, the prior and posterior models are trained employing a \textit{variational inference} approach, facilitating the identification of pivotal steps for the decoder's operation. This decoder employs \textit{progressive masks}, thereby engendering the realization of \textit{partially and progressive forecasting} strategies. The intricate implementation intricacies of these components are elaborated upon in the subsequent sections.
}
\subsection{Partial Autoregressive Forecasting}
Our ProNet makes predictions by combining AR and NAR decoding mechanisms together. 
We divide the steps in forecasting horizon into $n_g$ segments and predict the steps at starting positions of each segment $S_1=[s_1,s_2,...,s_{n_g}]$ simultaneously, and make predictions for their next steps at position $S_2=[s_1+1,s_2+1,...,s_{n_g}+1]$ autoregressively in parallel until all forecasts in horizon are made. Note that the starting position of the first segment is set as 1 (the first step). 
Hence we can calculate length of the $i$th segment as the difference between the starting position of two sequent segments: $T_i=s_{i+1}-s_i$ where $s_{g+1}=T_h$. Following NAR forecasting models, we set the input y of decoder for the first predictions as 0 because previous predictions have not been made. To forecast all steps in horizon, ProNet needs to make AR predictions for $n_{step}={max}(T_{i:n_g})$ times, which is the maximum segment length.   
For an instance, the decoder predicts $[y_{s_1},y_{s_2},...,y_{s_{n_g}}]$ as the first step and uses them to predict $[y_{s_1+1},y_{s_2+1},...,y_{s_{n_g+1}}]$ as the next step, and so on. This process is repeated until as predictions are made. 

Different from AR and NAR models, our probability distribution is modelled by:
\begin{equation}
\begin{split}
	&p\left(\mathbf{Y}_{i,T_l+1:T} \mid \mathbf{Y}_{i,1:T_l}, \mathbf{X}_{i, 1: T}\right)\\ 
	= &\prod_{t=1}^{T_{n_g}} \prod_{j=1}^{n_g} p\left({y}_{i, t}^j \mid 
	\mathbf{Y}_{i,1:T_l}, \mathbf{X}_{i, 1: T_l}, \right. \\&
	\left.
	\mathbf{Y}_{i, T_l+1: T_l+t}^1, \mathbf{X}_{i, T_l+1: T_l+t}^1,...\right. 
	\left.
	\mathbf{Y}_{i, T_l+1: T_l+t}^{n_g}, \mathbf{X}_{i, T_l+1: T_l+t}^{n_g}\right),
\end{split}
\end{equation}
where ${y}_{i, t}^j $ is prediction at $t$th step of the $j$th segment and $\mathbf{Y}_{i, T_l+1: T_l+t}^{j}$ denotes the prediction history up to step $t$ of the $j$th segment.

\subsection{Progressive Prediction}
ProNet divides the forecasting horizon into segments that may have different lengths. However, the number of AR steps is determined by the maximum segment length so that the predictions for some segments could be made before the AR iteration (line 3 to 8 of Algorithm \ref{ProMask}) ends. We let these completed segments continue forecasting the steps within their next segments which were already predicted before. This is motivated by the fact that inaccurate predictions could be made at the early steps of each segment because there is none or limited dependency information available at the beginning. 
Thus, we employ a progressive prediction strategy to re-predict the steps when more dependency information is available.

\subsection{Progressive Mask}

For AR Transformer decoder \cite{Transformer17NIPS}, its attention mask is a lower triangular matrix that prevents future information leakage; while for NAR Transformer decoder (e.g. Informer \cite{Informer21}), its attention is unmasked.  
However, these masking mechanisms are not suitable to ProNet because it is partially AR and leverage future information for prediction. Thus, we design the progressive masking mechanism to access the first $t$ steps of all segments for the $t$th step prediction. 

Given the sample size $N$, forecasting horizon length $T_h$ and segment size $n_g$, the progressive mask $M$ is created by Algorithm \ref{ProMask}. Initially, we take the top $n_g$ indexes of latent variable $z$ that encodes the importance of steps for forecasting and stores them as $ind$, which is also the starting position $S_1$. Then we set the elements of zero vector $row$ located at $ind$ as one. 
We iterate from 1 to the maximum AR step $n_step$ to create the mask $M$: firstly, we set the rows of mask $M$ that is located at $ind$ as the variable $row$; secondly, we increment all elements of $ind$ by one and limit their value by the upper bound of forecasting horizon $T_h$ as shown in line 5 and 6 respectively; thirdly, we update the elements of $row$ located at $ind$ as one. 

For instance, Fig. \ref{CreationProMask} illustrates how the elements change in Algorithm \ref{ProMask} from initialization to the final settings. 
We firstly initialize the mask $M$ as a $7\times 7$ zero matrix. For the first iteration, the starting position or the index is $ind=S_1=[1,3,5]$, which means ProNet predicts the 1st, 3rd and 5th steps simultaneously. Then, we update the temporary variable $row\rightarrow[1~0~1~0~1~0~0]$ (line 2 of Algorithm \ref{ProMask}) and use it to fill the 1st, 3rd and 5th row of $M$ (line 4 of Algorithm \ref{ProMask}) as shown in the upper right of Fig. 2. Afterwards, we increment elements of $ind\rightarrow[2,4,6]$ by one and update temporary variable $row\rightarrow[1~1~1~1~1~1~0]$. The second iteration is as the first one, while final iteration implements progressive prediction: we now have the variable $row\rightarrow[1~1~1~1~1~1~1]$ and index $ind=[3,6,7]$. We fill the 3th, 6th and 7th row of $M$ with $row$, which means we use all previous predictions to forecast the 7th step and re-forecast the 3th and 6th steps. 

\begin{figure}[!t]
\centering
$$
\left[\begin{array}{lllllll}
	0 & 0 & 0 & 0 & 0 & 0 & 0 \\
	0 & 0 & 0 & 0 & 0 & 0 & 0 \\
	0 & 0 & 0 & 0 & 0 & 0 & 0 \\
	0 & 0 & 0 & 0 & 0 & 0 & 0 \\
	0 & 0 & 0 & 0 & 0 & 0 & 0 \\
	0 & 0 & 0 & 0 & 0 & 0 & 0 \\
	0 & 0 & 0 & 0 & 0 & 0 & 0
\end{array}\right] \quad\left[\begin{array}{lllllll}
	\textbf{1} & 0 & \textbf{1} & 0 & \textbf{1} & 0 & 0 \\
	0 & 0 & 0 & 0 & 0 & 0 & 0 \\
	\textbf{1} & 0 & \textbf{1} & 0 & \textbf{1} & 0 & 0 \\
	0 & 0 & 0 & 0 & 0 & 0 & 0 \\
	\textbf{1} & 0 & \textbf{1} & 0 & \textbf{1} & 0 & 0 \\
	0 & 0 & 0 & 0 & 0 & 0 & 0 \\
	0 & 0 & 0 & 0 & 0 & 0 & 0
\end{array}\right]
$$
$$	
\left[\begin{array}{lllllll}
	1 & 0 & 1 & 0 & 1 & 0 & 0 \\
	\textbf{1} & \textbf{1}& \textbf{1} & \textbf{1} & \textbf{1} & \textbf{1} & 0 \\
	1 & 0 & 1 & 0 & 1 & 0 & 0 \\
	\textbf{1} & \textbf{1}& \textbf{1} & \textbf{1} & \textbf{1} & \textbf{1} & 0 \\
	1 & 0 & 1 & 0 & 1 & 0 & 0 \\
	\textbf{1} & \textbf{1}& \textbf{1} & \textbf{1} & \textbf{1} & \textbf{1} & 0 \\
	0 & 0 & 0 & 0 & 0 & 0 & 0
\end{array}\right] \quad\left[\begin{array}{lllllll}
	1 & 0 & 1 & 0 & 1 & 0 & 0 \\
	1 & 1 & 1 & 1 & 1 & 1 & 0 \\
	{1} & \textbf{1} & {1} & \textbf{1} & {1} & {1} & \textbf{1}\\
	1 & 1 & 1 & 1 & 1 & 1 & 0 \\
	{1} & \textbf{1} & {1} & \textbf{1} & {1} & {1} & \textbf{1}\\
	1 & 1 & 1 & 1 & 1 & 1 & 0 \\
	\textbf{1} & \textbf{1} & \textbf{1} & \textbf{1} & \textbf{1} & \textbf{1} & \textbf{1}
\end{array}\right]
$$
\caption{Creation process of progressive mask $M$: initial $M$ (upper left), $M$ after the 1st (upper right) and 2nd (lower left) iteration, and the final $M$ (lower right) when the forecasting horizon $T_h=7$, the segment size $n_g=3$ and starting positions of each segments $S_1=[1,3,5]$. We mark their changes in bold.}
\label{CreationProMask}
\end{figure}

\begin{algorithm}[t]
\caption{Creation of Progressive Mask}
\textbf{Input}:\hspace*{1pt} progressive mask $M\in\Re^{N\times T_h\times T_h}=\textbf{0}$, latent factor $z\in\Re^{N\times T_h}$, temporary variable $row\in\Re^{T_h}=\textbf{0}$.\\
\textbf{Output}:\hspace*{1pt}   progressive mask $M$.
\begin{algorithmic}[1]
	\STATE $ind=argtopk(z)$	;		
	\STATE $row[ind]=1$;
	
	\FOR{$1 \rightarrow n_{step}$:}
	\STATE $M[ind]=row$;
	\STATE $ind+=1$;
	\STATE $ind[ind>T_h]=T_h$;
	\STATE $row[ind]=1$;
	\ENDFOR 
\end{algorithmic}
\label{ProMask}
\end{algorithm}

\subsection{Variational Inference}

\begin{figure}[!t]
\centering	  
\subfigure[]{\includegraphics[width=.5\columnwidth]{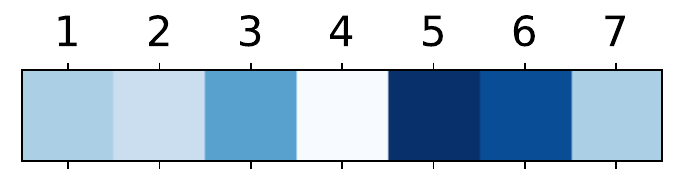}}		\subfigure[]{\includegraphics[width=.5\columnwidth]{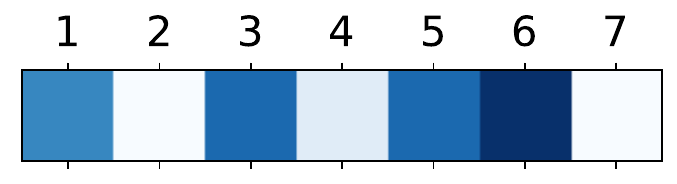}}	
\caption{Visualization of latent variable $z$: (a) original $z$, (b) re-weighted $z$. Higher brightness indicates the higher value of $z$ element. }
\label{zw}
\end{figure}

{
The ProNet algorithm addresses the challenge of segmenting sequences and prioritizing forecasted steps to achieve optimal performance. It is crucial to initiate forecasting with steps that carry the most significance and intricate dependencies on subsequent time points. However, obtaining this vital information is not straightforward. Drawing inspiration from the methodology introduced in \cite{FastDec18ICML}, we tackle this issue by employing parallel forecasting of step importance, representing them as latent variables denoted as $z$. These latent variables are derived through conditional variational inference, an approach rooted in conditional Variational Autoencoders (cVAEs) \cite{cVAE}. These cVAEs bridge the gap between observed and latent variables, facilitating a deeper understanding of data patterns.

The concept of cVAEs extends the classical Variational Autoencoder (VAE) framework \cite{Kingma2014}, enhancing it by integrating conditioning variables into the data generation process. This augmentation empowers cVAEs to learn a more nuanced and contextually aware latent space representation of data. In a standard VAE, data is mapped to a lower-dimensional latent space using an encoder, and subsequently, a decoder reconstructs this data from points in the latent space. cVAEs introduce conditional variables that encode additional context or prior knowledge into the generative model. This enables cVAEs not only to learn conditional latent representations but also to incorporate provided contextual cues effectively. Particularly, cVAEs are advantageous in scenarios where supplementary information is available, mirroring the case of ProNet, which requires generating initial time steps for predictions based on past ground truth and covariates.

In the context of ProNet, the latent variables, denoted as $z$, correspond to individual output steps and rely on the entire temporal sequence for their determination. Consequently, the conditional probability is articulated as:

\begin{equation}
	\begin{aligned}
	P_{\theta}({y}\mid{x})=	\int_{z}P_{\theta}({y} \mid z,{x})P_{\theta}(z\mid{x})dz
	\end{aligned}
\end{equation}
The $y$ denotes the ground truth in the forecasting horizon, and conditioning variable $x$ plays the role of historical data and covariates, allowing the model to capture the relevance of different time steps as latent variable $z$ for accurate predictions.

However, the direct optimization of this objective is unfeasible. To address it, the Evidence Lower Bound (ELBO) \cite{cVAE} is employed as the optimization target, resulting in the following formulation:
\begin{equation}
	\begin{aligned}
		\log P_{\theta}({y} \mid {x}) & \geq \mathrm{E}{q{\phi}({z} \mid {y}, {x})}\left[\log P_{\theta}({y} \mid {z}, {x})\right] \\
		&-\mathrm{KL}\left(q_{\phi}({z} \mid {y}, {x}) | p_{\theta}({z} \mid {x})\right)
	\end{aligned}
\end{equation}
Here, the Kullback-Leibler (KL) divergence is denoted by $\mathrm{KL}$. The term $p_{\theta}({z} \mid {x})$ represents the prior distribution, $q_{\phi}({z} \mid {y}, {x})$ denotes an approximated posterior, and $P_{\theta}({y} \mid {z}, {x})$ characterizes the decoder. With the ground truth encompassed within the horizon denoted by $y$ and the condition $x$, $q_{\phi}({z} \mid {y}, {x})$ effectively models the significance of diverse time steps represented by $z$. Notably, during prediction, $y$ is not available, prompting the need to train $p_{\theta}({z} \mid {x})$ to approximate $q_{\phi}({z} \mid {y}, {x})$, achieved through the minimization of the KL divergence. 

Both the prior and the approximated posterior are modelled as Gaussian distributions characterized by their mean and variance. The mean $\mu$ is obtained via a linear layer, while the variance $\sigma$ is derived through another linear layer, followed by a SoftPlus activation function. To enable smooth gradient flow through random nodes, the reparameterization trick \cite{cVAE} is invoked. This involves sampling the latent variable $z$ using the equation $z = g(\epsilon, \mu, \sigma) = \mu + \sigma \epsilon$, where $\epsilon$ follows a standard normal distribution $\mathcal{N}(0,1)$, effectively serving as white noise. The value of $z$ encapsulates the significance of each time step within the forecasting horizon, guiding the selection of which steps to initiate predictions from. The top $k$ indices featuring the highest $z$ values are chosen to initiate forecasting.

During the training process, $z$ is sampled from $q_{\phi}({z} \mid {y}, {x})$, and the approximation of $q_{\phi}({z} \mid {y}, {x})$ to the true posterior $p_{\theta}({z} \mid {x})$ is enforced. This entire framework enables ProNet to identify and leverage the most crucial time steps for accurate and effective forecasting.

Empirically, we find  both the prior and posterior models often assign elevated importance to a sequence of steps, leading to a substantial reduction in decoding speed during testing. Striking a balance between accuracy and speed, we introduce a novel approach to realign the latent factor $z$ by incorporating a scaling factor with the assistance of weight vectors denoted as $W\in\Re^{T_h-1}$:
\begin{equation}
\begin{aligned}
	z&=softmax(z)\times W\\
	W&=|cos([0,1,...,T_h-1]\times\frac{n_g\pi}{T_h})|
\end{aligned}
\end{equation}

This re-weighting operation modifies the latent factor $z$ to achieve a more optimized equilibrium between the forecasting accuracy and the computational speed.
Subsequently, we determine the initial position, denoted as $S_1$, and identify the indices of the largest $n_g-1$ elements from $z[2:]$ as potential starting positions. For example, Fig. \ref{zw} provides a visual representation of the latent variable $z$ before and after the re-weighting process. By selecting $n_g=3$, the original $z$ yields the starting position $S_1=[1,5,6]$, necessitating 4 autoregressive (AR) iterations to complete the forecasting process. Conversely, the re-weighted $z$ results in a starting position $S_1=[1,3,6]$, reducing the AR iterations required to 3. Remarkably, this re-weighting design elevates decoding speed by 25\% in this scenario.

Illustrating the tangible benefits of our approach, this strategic re-weighting of the latent variable $z$ not only preserves forecast accuracy but also significantly enhances the computational efficiency of the process.

}

\section{Experiments}

\begin{table*}[t]
\renewcommand{\arraystretch}{1}	
\centering
\begin{tabular}{C{1.5cm} C{1.8cm} C{1.8cm} C{2cm} C{.8cm} C{.8cm} C{.8cm} C{.8cm} C{.8cm} C{.8cm} } 
	\specialrule{.1em}{.05em}{.05em} 
	&Start date&End date &Granularity &$L_d$&$N$& $n_{T}$  & $n_{C}$& $T_{l}$  & $T_{h}$ \\	
	\hline
	Sanyo&01/01/2011& 31/12/2016 & 30 minutes&20 & 1 &4 &3&20 &20\\				
	Hanergy&01/01/2011& 31/12/2017 & 30 minutes&20 & 1 &4 &3&20 &20\\			
	Solar&01/01/2006& 31/08/2006 & 1 hour&24 & 137 &0 &3&24 &24\\			
	Electricity&01/01/2011& 07/09/2014 & 1 hour&24 & 370 &0 &4&168 &24\\		
	\specialrule{.1em}{.05em}{.05em} 
\end{tabular}
\caption{Dataset statistics. $L_d$ - number of steps per day, $N$ - number of series, $n_{T}$ - number of time-based features, $n_{C}$ - number of calendar features, $T_{l}$ - length of input series, $T_{h}$ - length of forecasting horizon. }
\label{datasets} 
\end{table*}

\subsection{Data Sets}
We conduct experiments on four publicly available time series datasets: Sanyo \cite{San}, Hanergy \cite{Han}, Solar \cite{Sol} and Electricity \cite{Ele}. They include: \textbf{Sanyo} and \textbf{Hanergy} - solar power generation data from two Australian PV plants for 6 and 7 years; \textbf{Solar} - solar power data from 137 US PV plants for 8 months, \textbf{Electricity} - electricity consumption data of 370 households for about 4 years. The covariate time series are weather, weather forecasts and calendar features for Sanyo and Hanergy and calendar features only for the other datasets. The data is sampled every 30 minutes for Sanyo and Hanergy sets and 60 minutes for Solar and Electricity sets. 
The data statistics is summarised in Table. \ref{datasets}.

For Sanyo and Hanergy, the data between 7am and 5pm is considered and was aggregated at half-hourly intervals. For both datasets, weather and weather forecast data was also collected (see \cite{Yang20ICONIP} for more details) and used as covariates. 
The Solar and Electricity data is aggregated into 1-hour intervals. 
Following \cite{Logsparse19NIPS,Yang20ICONIP}, calendar features are added according to the granularity of the datasets: Sanyo and Hanergy use \textit{month, hour-of-the-day and minute-of-the-hour}, Solar uses \textit{month, hour-of-the-day and age} and Electricity uses \textit{month, day-of-the-week, hour-of-the-day and age}. 

All data was normalized to have zero mean and unit variance. 

\subsection{Experimental Details}

\begin{table*}[t]
\renewcommand{\arraystretch}{1}	
\centering
\begin{tabular}{C{1.5cm}  C{.8cm} C{.8cm} C{.8cm} C{.8cm}  C{1cm} C{.8cm} C{.8cm} } 
	\specialrule{.1em}{.05em}{.05em} 
	&$\lambda$& $\delta$ & $d_{hid}$ & $n_{e}$& $n_{d}$ & $d_{f}$ &$n_{h}$\\	
	\hline
	Sanyo&0.005& 0.1 & 24 &3& 3 &32&4\\	
	Hanergy&0.005&0.1 & 24 & 2&2&32 &12 \\	
	Solar&0.005& 0.1& 48 & 4&3 &24 &12\\	
	Electricity&0.001& 0.1 & 48 &3&3 &32 &12\\
	\specialrule{.1em}{.05em}{.05em} 
\end{tabular}
\caption{Hyperparameters for ProNet}
\label{ProNet_Parameters} 
\end{table*}
We compare the performance of ProNet with seven methods: five state-of-the-art deep learning (DeepAR, DeepSSM, LogSparse Transformer, N-BEATS and Informer), a statistical (SARIMAX) and a persistence model:

\begin{itemize}
\item Persistence is a typical baseline in forecasting and considers the time series of the previous day as the prediction for the next day.
\item SARIMAX \cite{Durbin01book} is an extension of the ARIMA and can handle seasonality with exogenous factors.
\item DeepAR \cite{DeepAR20} is a widely used sequence-to-sequence probabilistic forecasting model. 
\item DeepSSM \cite{DeepSSM18NIPS} fuses SSM with RNNs to incorporate structural assumptions and learn complex patterns from the time series. It is the state-of-the-art deep forecasting model that employs SSM.
\item N-BEATS \cite{N-BEATS} consists of blocks of fully-connected neural networks, organised into stacks using residual links. We introduced covariates at the input of each block to facilitate multivariate series forecasting.
\item LogSparse Transformer \cite{Logsparse19NIPS} is a recently proposed variation of the Transformer architecture for time series forecasting with convolutional attention and sparse attention; it is denoted as "LogTrans" in Table \ref{Accuracy}. 
\item Informer \cite{Informer21} is a Transformer-based forecasting model based on the ProbSparse self-attention and self-attention distilling. 
We modified it for probabilistic forecasts to generate the mean value and variance. 
\end{itemize}
Note that Persistence, N-BEATS and Informer are NAR models while the others are AR models.

All models were implemented using PyTorch 1.6 and evaluated on Tesla V100 16GB GPU. 
The deep learning models were optimised by mini-batch gradient descent with the Adam optimiser and a maximum number of epochs 200. 

{
In line with the experimental setup from \cite{Yang20ICONIP} and \cite{Logsparse19NIPS}, we carefully partitioned the data to prevent future leakage during our evaluations. Specifically, for Sanyo and Hanergy datasets, we designated the data from the last year as the test set, the second last year as the validation set for early stopping, and the remaining data (5 years for Sanyo and 4 years for Hanergy) as the training set.
For the Solar and Electricity datasets, we utilized the data from the last week (starting from 25/08/2006 for Solar and 01/09/2014 for Electricity) as the test set, and the week preceding it as the validation set. To ensure consistency, the data preceding the validation set was further divided into three subsets, and the corresponding validation set was employed to select the best hyperparameters.
Throughout the process, our hyperparameter selection was based on achieving the minimum loss on the validation set, enabling us to fine-tune the model for optimal performance.}

We used Bayesian optimization for hyperparameter search for all deep learning models with a maximum number of iterations 20. The models used for comparison were tuned based on the recommendations in the original papers. We selected the hyperparameters with a minimum loss on the validation set. Probabilistic forecasting models use NLL loss while the point forecasting model(N-BEATS) uses mean squared loss.

For the Transformer-based models, we used learnable position and ID (for Solar and Electricity sets) embedding. 
For ProNet, the constant sampling factor for Informer backbone was set to 2, and the length of start token $T_de$ is fixed as half of the forecasting horizon. The learning rate $\lambda$ was fixed; the number of segments $n_g$ was fixed as 10 for Sanyo and Hanergy data sets, and 12 for Solar and Electricity sets; the dropout rate $\delta$ was chosen from \{0, 0.1, 0.2\}. The hidden layer dimension size $d_{hid}$ was chosen from \{8, 12, 16, 24, 32, 48, 96\}; the Informer backbone Pos-wise FFN dimension size $d_{f}$ and number of heads $n_{h}$ were chosen from \{8, 12, 16, 24, 32, 48, 96\} and \{4, 8, 16, 24, 32\}; the number of hidden layers of encoder $n_e$ and decoder $n_{d}$  were chosen from \{2, 3, 4\}. Following \cite{Informer21,kasai2021deep}, we restrict the decoder layers to be less than encoder layers for a fast decoding speed. 
The selected best hyperparameters for ProNet are listed in Table \ref{ProNet_Parameters} and used for the evaluation of the test set.

As in \cite{DeepAR20}, we report the standard $\rho$0.5 and $\rho$0.9-quantile losses. 
Note that $\rho$0.5 is equivalent to MAPE.  
Given the ground truth $y$ and $\rho$-quantile of the predicted distribution $\hat{y}$, the $\rho$-quantile loss is defined by: 
\begin{equation}
\begin{split}
	\mathrm{QL}_{\rho}(y, \hat{y})&=\frac{2\times\sum_{t} P_{\rho}\left(y_{t}, \hat{y}_{t}\right)}{\sum_{t}\left|y_{t}\right|}, 
	\\
	\quad P_{\rho}(y, \hat{y})&=\left\{\begin{array}{ll}
		\rho(y-\hat{y}) & \text { if } y>\hat{y} \\
		(1-\rho)(\hat{y}-y) & \text { otherwise }
	\end{array}\right.
\end{split}
\end{equation}

\begin{table*}[t]
\renewcommand{\arraystretch}{1}	
\centering
	\begin{tabular}{  C{2.7cm}   C{2.2cm}  C{2.2cm}  C{2.2cm}  C{2.2cm}  } 
		\specialrule{.1em}{.05em}{.05em} 
		& Sanyo& Hanery& Solar& Electricity\\
		\hline
		Persistence &0.154/-&0.242/-&0.256/-&0.091/-\\ 
		SARIMAX&0.124/0.096&0.145/0.098&0.256/0.192&0.196/0.079\\
		DeepAR 
		&0.070/0.031&0.092/0.045&0.222$^\diamond$/0.093$^\diamond$&0.075$^\diamond$/0.040$^\diamond$\\	
		DeepSSM
		&\textbf{0.042}/0.023&\textbf{0.070}/0.053&0.223/0.181&0.083$^\diamond$/0.056$^\diamond$\\	
		LogTrans
		&0.067/0.036&0.124/0.066&0.210$^\diamond$/\textbf{0.082}$^\diamond$&\textbf{0.059}$^\diamond$/{0.034}$^\diamond$\\
		N-BEATS&0.077/-&0.132/-&0.212/-&0.071/-\\			
		Informer&{0.046}/{0.022}&{0.084}/{0.046}&{0.215}/{0.115}&0.068/0.033\\
		ProNet&\textbf{0.042}/\textbf{0.021}&\textbf{0.070}/\textbf{0.035}&\textbf{0.205}/{0.091}&{0.071}/\textbf{0.32}\\	
		\specialrule{.1em}{.05em}{.05em} 
	\end{tabular}
	\caption{$\rho$0.5/$\rho$0.9-loss of data sets with various granularities. $\diamond$ denotes results from \protect\cite{Logsparse19NIPS}.}
	\label{Accuracy} 
\end{table*}

\section{Results}

\subsection{Accuracy Analysis}

Table \ref{Accuracy} shows the $\rho$0.5 and $\rho$0.9 losses of all models. As N-BEATS and Persistence produce point forecasts, only the $\rho$0.5-loss is reported for them. 

We can see ProNet is the most accurate method - it outperforms other methods on all data sets except for $\rho$0.9 on Solar and $\rho$0.5 on Electricity where Logsparse Transformer shows better performance. A possible explanation is that the ProNet backbone - Informer has subpar performance for the two cases. As a NAR forecasting model, Informer ignores dependency in target space, while our ProNet assumes the alternative dependency and therefore achieves better accuracy than Informer.   
Comparing the performance of AR and NAR models, we can see our ProNet is the most successful overall - ProNet achieves a trade-off between AR and NAR forecasting models by assuming an alternative dependency and accessing both past and future information for forecasting with latent variables.

\subsection{Visualization Analysis}
\begin{figure}[!ht]
	\centering	  
	\subfigure[]{\includegraphics[width=0.6\textwidth]{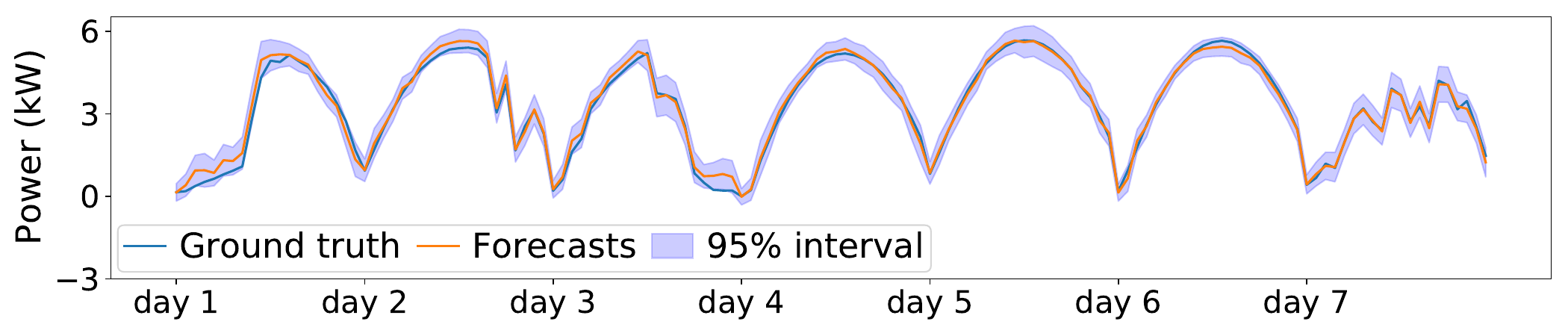}}		\subfigure[]{\includegraphics[width=.6\textwidth]{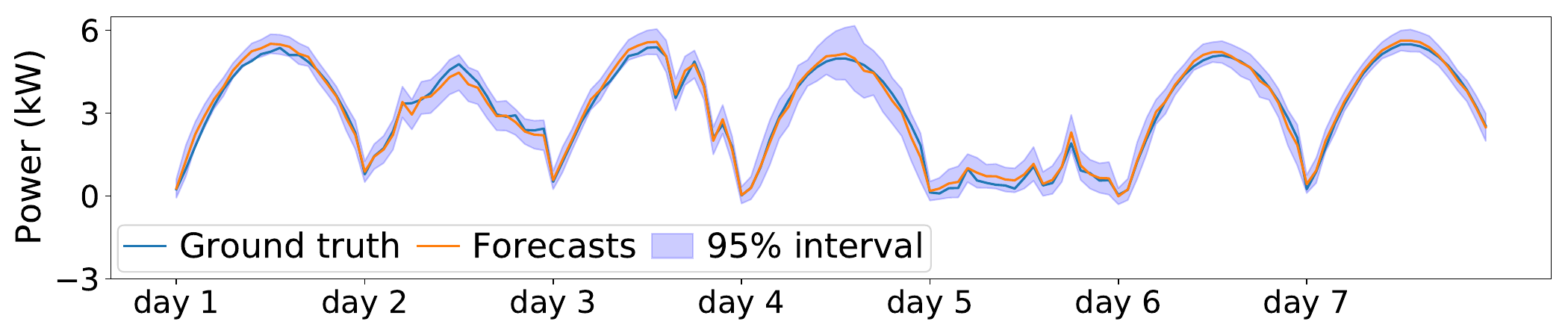}}	
	\subfigure[]{\includegraphics[width=.6\textwidth]{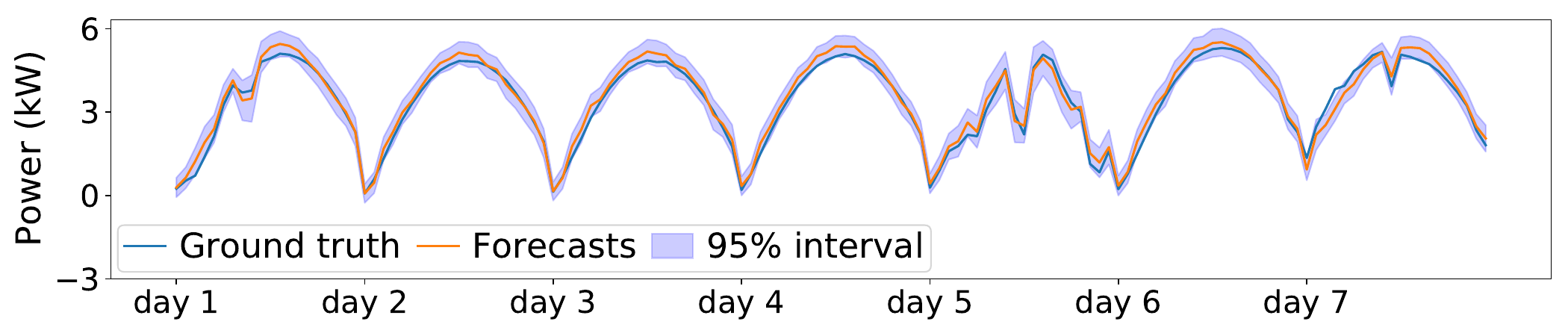}}		\subfigure[]{\includegraphics[width=.6\textwidth]{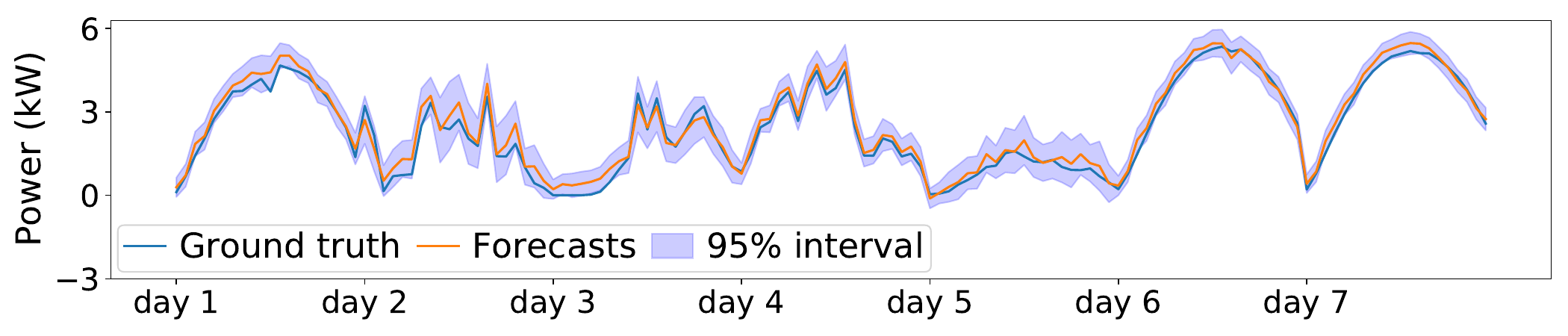}}	
	\subfigure[]{\includegraphics[width=.6\textwidth]{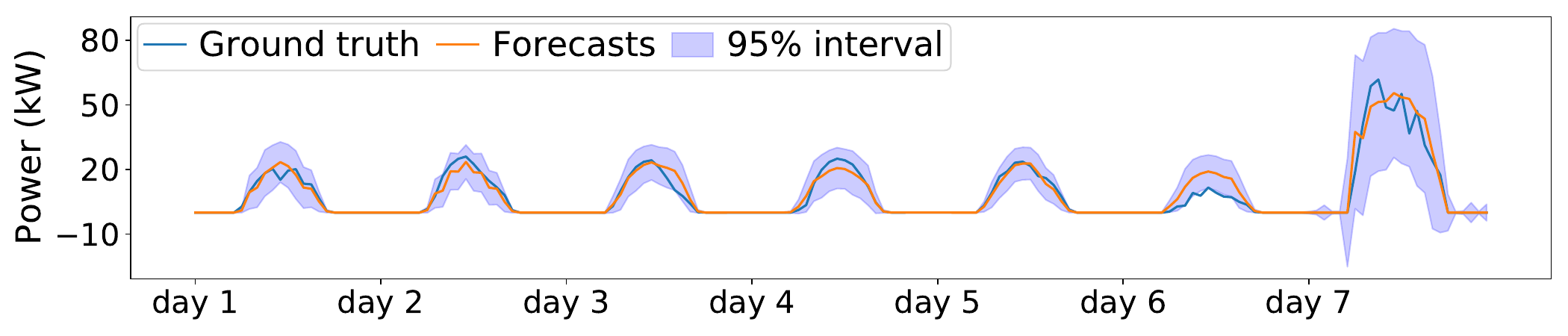}}			\subfigure[]{\includegraphics[width=.6\textwidth]{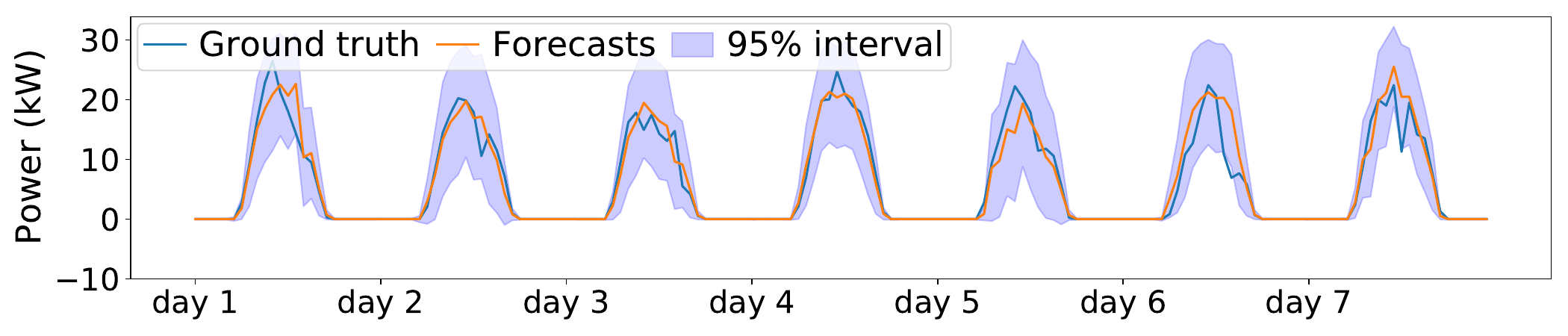}}		
	\caption{Actual vs ProNet predicted data with trend and seasonality components and 95$\%$ confidence  intervals: (a) and (b) - Sanyo; (c) and (d) - Hanergy; (e) and (f) - Solar data sets.
	}
	\label{Visualized_results}
\end{figure}

\begin{figure}[!ht]
	\centering	  	
	\subfigure[]{\includegraphics[width=.6\columnwidth]{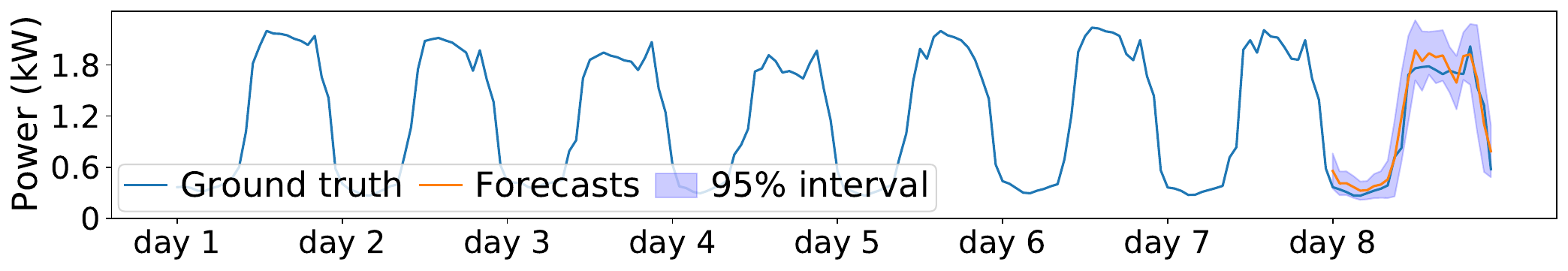}}	\subfigure[]{\includegraphics[width=.6\columnwidth]{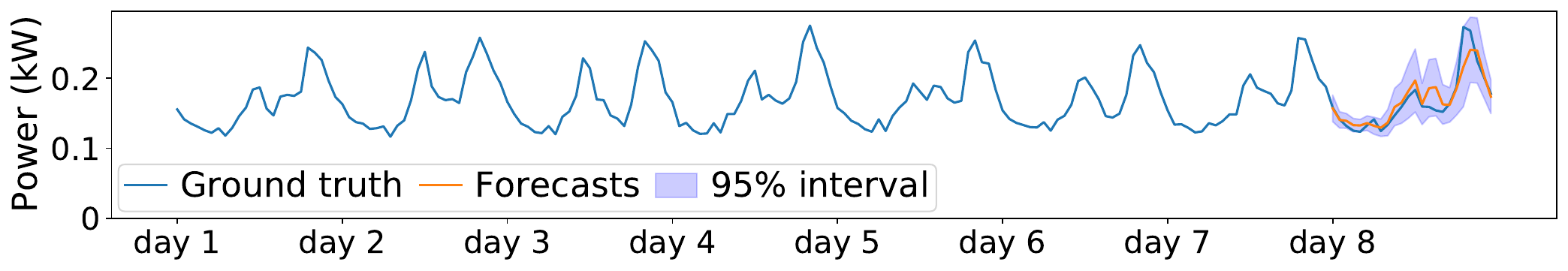}}
	\caption{ProNet case study on Electricity data set: actual vs predicted data.}
	\label{elect_pre}
\end{figure}

Fig. \ref{Visualized_results} illustrates example forecasts from our ProNet on Sanyo, Hanergy and Solar sets. We can see ProNet can make accurate forecasts and capture the complex and varying patterns in the forecasting horizon. 
Fig. \ref{elect_pre} presents the results for 8 consecutive days from the test set for the Electricity data: 7-day past history and 1-day forecasting output of SSDNet. It shows ProNet can learn from related series and long past history to predict accurately.


\subsection{Error Accumulation}

To investigate the ability of ProNet to handle error accumulation and model the output distribution, we compare ProNet with an AR model (DeepAR) and a NAR model (Informer) on the Sanyo and Hanergy as a case study. 

\begin{figure}[ht]
	\centering	  
	\subfigure[]{\includegraphics[width=.48\columnwidth]{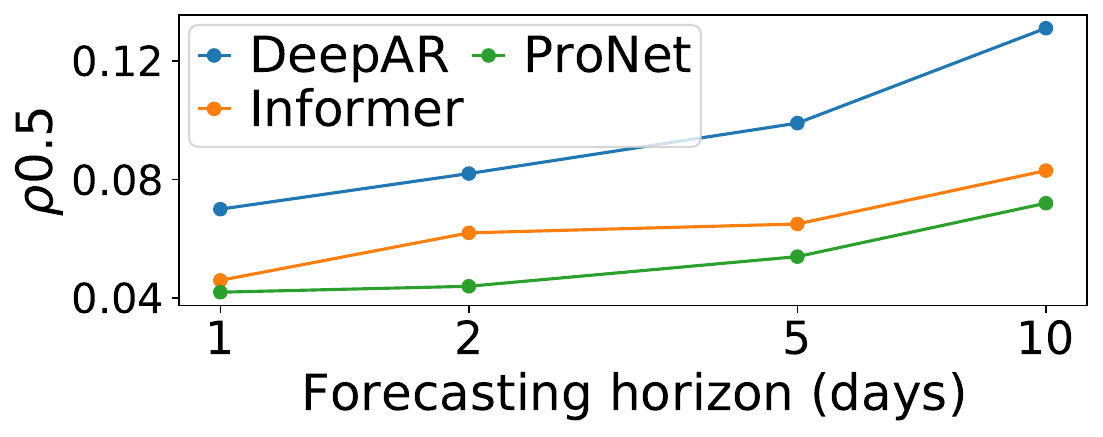}}
	\subfigure[]{\includegraphics[width=.48\columnwidth]{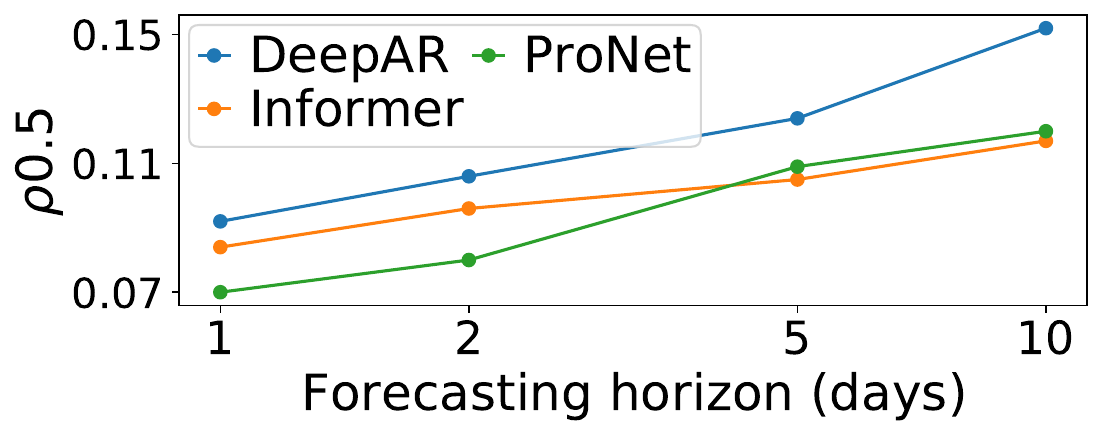}}	
	\caption{$\rho$0.5-loss 
		for various forecasting horizons on (a) Sanyo and (b) Hanergy datasets.}
	\label{Error_multiH}
\end{figure}

Fig. \ref{Error_multiH} shows the $\rho$0.5-loss of the models for the forecasting horizons range from 1 (20 steps) to 10 days (200 steps).
We can see the $\rho$0.5-loss of all models increases with the forecasting horizon but the performance of DeepAR drops more significantly due to its AR decoding mechanism and error accumulation. ProNet consistently outperforms Informer for short horizon and has competitive performance with Informer for long horizon, indicating the effectiveness of seeking the trade-off between AR and NAR models. ProNet assumes the dependency in target space without fully discarding AR decoding and can improve the forecasting accuracy over all horizons.  

The results show that error accumulation degrades the performance of AR models but ProNet can successfully overcome this by assuming the alternative dependency and fusing future information into predictions with a shorter AR decoding path.

\subsection{Inference Speed}
We evaluate the prediction time of ProNet with varying number of segments $n_g$ and compare it with the AR and NAR model: LogTrans and Informer. 
Table \ref{Inference_time} shows the average elapsed time and the standard deviation over 10 runs; all models were run on the same computer configuration.  

As expected, ProNet has a faster inference speed than the AR LogTrans for their shorter AR decoding path. 
The inference speed of ProNet increases with the number of segments $n_g$ up to 10. This is because the number of AR steps decreases with $n_g$. ProNet with $n_g=10$ and $n_g=15$ have similar speed as both are expected to have same number of steps 2. 
As the number of segments $n_g$ increases, ProNet has competitive inference speed with Informer when  $n_g=10$ and $n_g=15$.

The results confirm that ProNet remains the fast decoding advantage of NAR models, in addition to being the most accurate.

\begin{table}[!ht]
	\renewcommand{\arraystretch}{1}	
	\centering
	\begin{tabular}{ C{3cm}   C{2cm}  C{2cm}   } 
		\specialrule{.1em}{.05em}{.05em} 
		& Sanyo& Hanergy\\
		\hline
		Informer&18.1$\pm$0.5&18.7$\pm$1.1\\
		LogTrans&101.5$\pm$3.5&112.7$\pm$7.7\\
		\hline
		ProNet ($n_g=2$)&{99.8}$\pm$6.5&{76.7}$\pm$8.8\\
		ProNet ($n_g=5$)&{52.3}$\pm$1.6&{43.8}$\pm$5.9\\
		ProNet ($n_g=10$)&{22.8}$\pm$0.1&{21.3}$\pm$0.2\\
		ProNet ($n_g=15$)&{21.6}$\pm$0.6&{20.4}$\pm$3.9\\
		\specialrule{.1em}{.05em}{.05em} 
	\end{tabular}
	\caption{Prediction time (ms) - mean and standard deviation.
	}
	\label{Inference_time} 
\end{table}

\subsection{Ablation and Hyperparameter Sensitivity Analysis}

To evaluate the effectiveness of proposed methods, we conducted an ablation study on Sanyo and Hanergy sets. Table \ref{Ablation_Accuracy} shows the performance of: 1) Trans (AR Transformer); 2) PAR-Trans is the partially AR Transformer implemented by simply dividing the horizon evenly \cite{Fei21AAAI}; 3) ProNet-Trans is the ProNet that uses Transformer as backbone instead of Informer; 4) Informer; 5) PAR-Informer is the partially AR Informer \cite{Fei21AAAI}; 6) our ProNet. 

We can see PAR-Trans outperforms Trans and PAR-Informer performs worse than Informer that indicate the partially AR decoding mechanism can improve Trans but degrades the performance of Informer. A possible explanation is that simply dividing forecasting horizon into even segments and the fixed dependency assumption violates the real data distribution, which has time-varying dependency relationships (see Fig. \ref{elect_pc}). Both ProNet-Trans and ProNet outperform Trans and Informer as well as their partially AR version consistently, showing the effectiveness of our progressive decoding mechanism and confirming the advantage of it over partially AR decoding mechanism.

\begin{table}[t]
	\renewcommand{\arraystretch}{1}	
	\centering
		\begin{tabular}{  C{4cm}   C{2.2cm}  C{2.2cm}   } 
			\specialrule{.1em}{.05em}{.05em} 
			& Sanyo& Hanery\\
			\hline	
			Trans&0.074/0.035&0.120/0.058\\ 			
			PAR-Trans&0.064/0.031&0.107/0.054\\ 	
			ProNet-w/o-Trans&0.064/0.031&0.107/0.054\\ 	
			ProNet-Trans&0.055/0.030&0.089/0.044\\ 
			\hline
			Informer&{0.046}/{0.022}&{0.084}/{0.046}\\
			PAR-Informer&{0.059}/{0.032}&{0.096}/{0.046}\\
			ProNet-w/o-Informer&0.064/0.031&0.107/0.054\\ 	
			ProNet&{0.042}/0.021&{0.070}/{0.035}\\	
			\specialrule{.1em}{.05em}{.05em} 
		\end{tabular}
		\caption{$\rho$0.5/$\rho$0.9-loss for ablation study.}
		\label{Ablation_Accuracy} 
	\end{table}
	
	We perform the sensitivity analysis of the proposed ProNet on Sanyo and Hanergy sets. 
	Table \ref{AccuracyN_g} shows the $\rho$0.5/$\rho$0.9-loss of ProNet with the number of segments $n_g$ ranges from 2 to 15. 
	ProNet achieves the optimal trade-off with 5 and 10 segments $n_g$, in which cases the performance is the best. It can be explained that when $n_g$ is low, more AR decoding steps are required and error accumulates; when  $n_g$ is high, most steps of ProNet are predicted non-autoregressively without the dependency in target space.  
	In summary, considering the ProNet inference speed as provided in Table \ref{Inference_time}, dividing the forecasting horizon by half is the best choice that allows ProNet to achieve the best accuracy and speed. 
	
	\begin{table*}[t]
		\renewcommand{\arraystretch}{1}	
		\centering
		
		\begin{tabular}{  C{1.4cm}  C{2.4cm}   C{2.4cm}  C{2.4cm}  C{2.4cm}  } 
			\specialrule{.1em}{.05em}{.05em} 
			$n_g$&2& 5& 10& 15\\
			\hline
			Sanyo &0.041/0.022&0.040/0.021&0.042/0.021&0.042/0.021\\ 
			Hanery&0.074/0.033&0.073/0.033&0.070/0.035&0.075/0.033\\ 
			\specialrule{.1em}{.05em}{.05em} 
		\end{tabular}
		\caption{$\rho$0.5/$\rho$0.9-loss of data sets with various number of segments $n_g$ for ProNet.}
		\label{AccuracyN_g} 
	\end{table*}
	\begin{table*}[t]
	\renewcommand{\arraystretch}{1}	
	\centering
	
	\begin{tabular}{  C{1.4cm}  C{2.4cm}   C{2.4cm}  C{2.4cm}  C{2.4cm}  } 
		\specialrule{.1em}{.05em}{.05em} 
		$n_g$&2& 5& 10& 15\\
		\hline
		Sanyo &0.040/0.022&0.045/0.023&0.047/0.024&0.041/0.022\\ 
		Hanery&0.075/0.031&0.072/0.033&0.076/0.039&0.078/0.035\\ 
		\specialrule{.1em}{.05em}{.05em} 
	\end{tabular}
	\caption{$\rho$0.5/$\rho$0.9-loss of data sets with various number of segments $n_g$ for ProNet without re-weighting mechanism.}
	\label{AccuracyN_gw/o} 
	\end{table*}

	\begin{table}[!ht]
		\renewcommand{\arraystretch}{1}	
		\centering
		\begin{tabular}{ C{3cm}   C{2cm}  C{2cm}   } 
			\specialrule{.1em}{.05em}{.05em} 
			& Sanyo& Hanergy\\
			\hline
			ProNet-w/o ($n_g=2$)&{123.2}$\pm$9.2&{95.2}$\pm$9.2\\
			ProNet-w/o ($n_g=5$)&{78.1}$\pm$5.1&{64.2}$\pm$7.5\\
			ProNet-w/o ($n_g=10$)&{59.3}$\pm$3.1&{52.1}$\pm$3.5\\
			ProNet-w/o ($n_g=15$)&{52.8}$\pm$2.2&{42.3}$\pm$3.5\\
			\specialrule{.1em}{.05em}{.05em} 
		\end{tabular}
		\caption{Prediction time (ms) ProNet without re-weighting mechanism - mean and standard deviation.
		}
		\label{Inference_timew/o} 
	\end{table}
	
	{
	Table \ref{AccuracyN_gw/o} and \ref{Inference_timew/o} present the evaluation of ProNet's $\rho$0.5/$\rho$0.9-loss performance and prediction speed without the re-weighting mechanism, across varying segment numbers ($n_s$). Comparing these results with the performance metrics of ProNet showcased in Table \ref{AccuracyN_g} and \ref{Inference_time}, it becomes evident that ProNet exhibits significantly higher prediction speeds when the re-weighting mechanism is absent. Furthermore, ProNet outperforms its re-weighting mechanism-less counterpart in 10 out of the 16 cases examined.
	
	This highlights the important role played by the re-weighting mechanism in enhancing ProNet's prediction speed while preserving its prediction accuracy. The incorporation of this mechanism effectively prevents the assignment of undue importance to specific sequences of steps, thus contributing to the optimization of prediction speed without compromising the overall accuracy of ProNet's forecasting.
	}

	\section{Conclusions}
	
{
We introduced ProNet, a novel deep learning model tailored for multi-horizon time series forecasting. ProNet effectively strikes a balance between autoregressive (AR) and non-autoregressive (NAR) models, avoiding error accumulation and slow prediction while maintaining the ability to model target step dependencies.
The key innovation of ProNet lies in its partially AR decoding mechanism, achieved through segmenting the forecasting horizon. It predicts a group of steps non-autoregressively within each segment while locally employing AR decoding, resulting in enhanced forecasting accuracy. The segmentation process relies on latent variables, effectively capturing the significance of steps in the horizon, optimized through variational inference. 
By embracing alternative dependency assumptions and fusing both past and future information, ProNet demonstrates its versatility and effectiveness in forecasting. Extensive experiments validate the superiority of our partially AR method, showcasing ProNet's remarkable performance and prediction speed compared to state-of-the-art AR and NAR forecasting models.

For future work, we envision further improvements by incorporating the latest decomposition and ensemble learning methods \cite{DU2022155,WANG2023122504} into ProNet, aiming to elevate the forecasting accuracy and robustness of our model.
}
	\bibliographystyle{cas-model2-names}
	\bibliography{Bibliography/bib_yang}

\begin{thebibliography}{50}
\expandafter\ifx\csname natexlab\endcsname\relax\def\natexlab#1{#1}\fi
\providecommand{\url}[1]{\texttt{#1}}
\providecommand{\href}[2]{#2}
\providecommand{\path}[1]{#1}
\providecommand{\DOIprefix}{doi:}
\providecommand{\ArXivprefix}{arXiv:}
\providecommand{\URLprefix}{URL: }
\providecommand{\Pubmedprefix}{pmid:}
\providecommand{\doi}[1]{\href{http://dx.doi.org/#1}{\path{#1}}}
\providecommand{\Pubmed}[1]{\href{pmid:#1}{\path{#1}}}
\providecommand{\bibinfo}[2]{#2}
\ifx\xfnm\relax \def\xfnm[#1]{\unskip,\space#1}\fi
\bibitem[{Barezi et~al.(2020)Barezi, Calixto, Cho and Fung}]{barezi2020study}
\bibinfo{author}{Barezi, E.J.}, \bibinfo{author}{Calixto, I.},
  \bibinfo{author}{Cho, K.}, \bibinfo{author}{Fung, P.}, \bibinfo{year}{2020}.
\newblock \bibinfo{title}{A study on the autoregressive and non-autoregressive
  multi-label learning}.
\newblock \href{http://arxiv.org/abs/2012.01711}{\tt arXiv:2012.01711}.
\bibitem[{Bengio et~al.(2015)Bengio, Vinyals, Jaitly and
  Shazeer}]{bengio2015scheduled}
\bibinfo{author}{Bengio, S.}, \bibinfo{author}{Vinyals, O.},
  \bibinfo{author}{Jaitly, N.}, \bibinfo{author}{Shazeer, N.},
  \bibinfo{year}{2015}.
\newblock \bibinfo{title}{Scheduled sampling for sequence prediction with
  recurrent neural networks}, in: \bibinfo{booktitle}{Proceedings of the
  Conference on Neural Information Processing Systems (NeurIPS)}.
\bibitem[{Chen and Lee(2015)}]{CHEN201599}
\bibinfo{author}{Chen, T.T.}, \bibinfo{author}{Lee, S.J.},
  \bibinfo{year}{2015}.
\newblock \bibinfo{title}{A weighted ls-svm based learning system for time
  series forecasting}.
\newblock \bibinfo{journal}{Information Sciences} \bibinfo{volume}{299},
  \bibinfo{pages}{99--116}.
\newblock \URLprefix
  \url{https://www.sciencedirect.com/science/article/pii/S0020025514011736},
  \DOIprefix\doi{https://doi.org/10.1016/j.ins.2014.12.031}.
\bibitem[{{D1}(2020)}]{San}
\bibinfo{author}{{D1}}, \bibinfo{year}{2020}.
\newblock \bibinfo{title}{Sanyo dataset}.
\newblock
  \bibinfo{howpublished}{http://dkasolarcentre.com.au/source/alice-springs/dka-m4-b-phase}.
\bibitem[{{D2}(2020)}]{Han}
\bibinfo{author}{{D2}}, \bibinfo{year}{2020}.
\newblock \bibinfo{title}{Hanergy dataset}.
\newblock
  \bibinfo{howpublished}{http://dkasolarcentre.com.au/source/alice-springs/dka-m16-b-phase}.
\bibitem[{{D3}(2014)}]{Sol}
\bibinfo{author}{{D3}}, \bibinfo{year}{2014}.
\newblock \bibinfo{title}{Solar dataset}.
\newblock
  \bibinfo{howpublished}{https://www.nrel.gov/grid/solar-power-data.html}.
\bibitem[{{D4}(2015)}]{Ele}
\bibinfo{author}{{D4}}, \bibinfo{year}{2015}.
\newblock \bibinfo{title}{Electricity dataset}.
\newblock
  \bibinfo{howpublished}{https://archive.ics.uci.edu/ml/datasets/ElectricityLoadDiagrams20112014}.
\bibitem[{Du et~al.(2022)Du, Gao, Suganthan and Wang}]{DU2022155}
\bibinfo{author}{Du, L.}, \bibinfo{author}{Gao, R.},
  \bibinfo{author}{Suganthan, P.N.}, \bibinfo{author}{Wang, D.Z.},
  \bibinfo{year}{2022}.
\newblock \bibinfo{title}{Bayesian optimization based dynamic ensemble for time
  series forecasting}.
\newblock \bibinfo{journal}{Information Sciences} \bibinfo{volume}{591},
  \bibinfo{pages}{155--175}.
\newblock \URLprefix
  \url{https://www.sciencedirect.com/science/article/pii/S0020025522000135},
  \DOIprefix\doi{https://doi.org/10.1016/j.ins.2022.01.010}.
\bibitem[{Durbin and Koopman(2001)}]{Durbin01book}
\bibinfo{author}{Durbin, J.}, \bibinfo{author}{Koopman, S.J.},
  \bibinfo{year}{2001}.
\newblock \bibinfo{title}{{Time Series Analysis by State Space Methods}}.
\newblock \bibinfo{publisher}{Oxford University Press}.
\bibitem[{Egrioglu et~al.(2022)Egrioglu, Baş and Chen}]{EGRIOGLU2022572}
\bibinfo{author}{Egrioglu, E.}, \bibinfo{author}{Baş, E.},
  \bibinfo{author}{Chen, M.Y.}, \bibinfo{year}{2022}.
\newblock \bibinfo{title}{Recurrent dendritic neuron model artificial neural
  network for time series forecasting}.
\newblock \bibinfo{journal}{Information Sciences} \bibinfo{volume}{607},
  \bibinfo{pages}{572--584}.
\newblock \URLprefix
  \url{https://www.sciencedirect.com/science/article/pii/S0020025522005941},
  \DOIprefix\doi{https://doi.org/10.1016/j.ins.2022.06.012}.
\bibitem[{Fei(2021)}]{Fei21AAAI}
\bibinfo{author}{Fei, Z.}, \bibinfo{year}{2021}.
\newblock \bibinfo{title}{Partially non-autoregressive image captioning}, in:
  \bibinfo{booktitle}{Proceedings of the AAAI Conference on Artificial
  Intelligence}.
\bibitem[{Gao et~al.(2022a)Gao, Wen and Deng}]{GAO2022553}
\bibinfo{author}{Gao, Q.}, \bibinfo{author}{Wen, T.}, \bibinfo{author}{Deng,
  Y.}, \bibinfo{year}{2022}a.
\newblock \bibinfo{title}{A novel network-based and divergence-based time
  series forecasting method}.
\newblock \bibinfo{journal}{Information Sciences} \bibinfo{volume}{612},
  \bibinfo{pages}{553--562}.
\newblock \URLprefix
  \url{https://www.sciencedirect.com/science/article/pii/S0020025522010404},
  \DOIprefix\doi{https://doi.org/10.1016/j.ins.2022.08.120}.
\bibitem[{Gao et~al.(2022b)Gao, Du, Suganthan, Zhou and Yuen}]{GAO2022117784}
\bibinfo{author}{Gao, R.}, \bibinfo{author}{Du, L.},
  \bibinfo{author}{Suganthan, P.N.}, \bibinfo{author}{Zhou, Q.},
  \bibinfo{author}{Yuen, K.F.}, \bibinfo{year}{2022}b.
\newblock \bibinfo{title}{Random vector functional link neural network based
  ensemble deep learning for short-term load forecasting}.
\newblock \bibinfo{journal}{Expert Systems with Applications}
  \bibinfo{volume}{206}, \bibinfo{pages}{117784}.
\bibitem[{Gao et~al.(2023)Gao, Li, Hu, Suganthan and Yuen}]{GAO202351}
\bibinfo{author}{Gao, R.}, \bibinfo{author}{Li, R.}, \bibinfo{author}{Hu, M.},
  \bibinfo{author}{Suganthan, P.}, \bibinfo{author}{Yuen, K.F.},
  \bibinfo{year}{2023}.
\newblock \bibinfo{title}{Online dynamic ensemble deep random vector functional
  link neural network for forecasting}.
\newblock \bibinfo{journal}{Neural Networks} \bibinfo{volume}{166},
  \bibinfo{pages}{51--69}.
\newblock \DOIprefix\doi{https://doi.org/10.1016/j.neunet.2023.06.042}.
\bibitem[{Gu et~al.(2018)Gu, Bradbury, Xiong, Li and Socher}]{Gu18ICLR}
\bibinfo{author}{Gu, J.}, \bibinfo{author}{Bradbury, J.},
  \bibinfo{author}{Xiong, C.}, \bibinfo{author}{Li, V.O.K.},
  \bibinfo{author}{Socher, R.}, \bibinfo{year}{2018}.
\newblock \bibinfo{title}{Non-autoregressive neural machine translation}, in:
  \bibinfo{booktitle}{Proceedings of the International Conference on Learning
  Representations (ICLR)}.
\bibitem[{Gu and Kong(2020)}]{Gu20Tricks}
\bibinfo{author}{Gu, J.}, \bibinfo{author}{Kong, X.}, \bibinfo{year}{2020}.
\newblock \bibinfo{title}{Fully non-autoregressive neural machine translation:
  Tricks of the trade}.
\newblock \bibinfo{journal}{arXiv preprint arXiv\: 2012.15833}
  \href{http://arxiv.org/abs/2012.15833}{\tt arXiv:2012.15833}.
\bibitem[{Han et~al.(2023)Han, Kumar and Tsvetkov}]{han-etal-2023-ssd}
\bibinfo{author}{Han, X.}, \bibinfo{author}{Kumar, S.},
  \bibinfo{author}{Tsvetkov, Y.}, \bibinfo{year}{2023}.
\newblock \bibinfo{title}{{SSD}-{LM}: Semi-autoregressive simplex-based
  diffusion language model for text generation and modular control}, in:
  \bibinfo{booktitle}{Proceedings of the 61st Annual Meeting of the Association
  for Computational Linguistics (Volume 1: Long Papers)},
  \bibinfo{publisher}{Association for Computational Linguistics},
  \bibinfo{address}{Toronto, Canada}. pp. \bibinfo{pages}{11575--11596}.
\newblock \URLprefix \url{https://aclanthology.org/2023.acl-long.647}.
\bibitem[{Hinton et~al.(2015)Hinton, Vinyals and Dean}]{KD15}
\bibinfo{author}{Hinton, G.}, \bibinfo{author}{Vinyals, O.},
  \bibinfo{author}{Dean, J.}, \bibinfo{year}{2015}.
\newblock \bibinfo{title}{Distilling the knowledge in a neural network}.
\newblock \bibinfo{journal}{arXiv preprint arXiv\: 1503.02531}
  \href{http://arxiv.org/abs/1503.02531}{\tt arXiv:1503.02531}.
\bibitem[{Hyndman and Athanasopoulos(2019)}]{Rob2019book}
\bibinfo{author}{Hyndman, R.J.}, \bibinfo{author}{Athanasopoulos, G.},
  \bibinfo{year}{2019}.
\newblock \bibinfo{title}{Forecasting: principles and practice}.
\newblock \bibinfo{publisher}{OTexts}.
\bibitem[{Kaiser et~al.(2018)Kaiser, Bengio, Roy, Vaswani, Parmar, Uszkoreit
  and Shazeer}]{FastDec18ICML}
\bibinfo{author}{Kaiser, L.}, \bibinfo{author}{Bengio, S.},
  \bibinfo{author}{Roy, A.}, \bibinfo{author}{Vaswani, A.},
  \bibinfo{author}{Parmar, N.}, \bibinfo{author}{Uszkoreit, J.},
  \bibinfo{author}{Shazeer, N.}, \bibinfo{year}{2018}.
\newblock \bibinfo{title}{Fast decoding in sequence models using discrete
  latent variables}, in: \bibinfo{booktitle}{Proceedings of the International
  Conference on Machine Learning (ICML)}.
\bibitem[{Kasai et~al.(2021)Kasai, Pappas, Peng, Cross and
  Smith}]{kasai2021deep}
\bibinfo{author}{Kasai, J.}, \bibinfo{author}{Pappas, N.},
  \bibinfo{author}{Peng, H.}, \bibinfo{author}{Cross, J.},
  \bibinfo{author}{Smith, N.A.}, \bibinfo{year}{2021}.
\newblock \bibinfo{title}{Deep encoder, shallow decoder: Reevaluating
  non-autoregressive machine translation}.
\newblock \bibinfo{journal}{CoRR} \href{http://arxiv.org/abs/2006.10369}{\tt
  arXiv:2006.10369}.
\bibitem[{Kim and Rush(2016)}]{Kim16EMNLP}
\bibinfo{author}{Kim, Y.}, \bibinfo{author}{Rush, A.M.}, \bibinfo{year}{2016}.
\newblock \bibinfo{title}{Sequence-level knowledge distillation}, in:
  \bibinfo{booktitle}{Proceedings of the Conference on Empirical Methods in
  Natural Language Processing (EMNLP)}.
\bibitem[{Kingma and Welling()}]{Kingma2014}
\bibinfo{author}{Kingma, D.P.}, \bibinfo{author}{Welling, M.}, .
\newblock in: \bibinfo{booktitle}{2nd International Conference on Learning
  Representations, {ICLR} 2014, timestamp = {2021-02-01T17:13:18.000+0100},
  title = {{Auto-Encoding Variational Bayes}}, year = 2014}.
\bibitem[{Li et~al.(2023)Li, Gao and Suganthan}]{LI2023833}
\bibinfo{author}{Li, R.}, \bibinfo{author}{Gao, R.},
  \bibinfo{author}{Suganthan, P.N.}, \bibinfo{year}{2023}.
\newblock \bibinfo{title}{A decomposition-based hybrid ensemble cnn framework
  for driver fatigue recognition}.
\newblock \bibinfo{journal}{Information Sciences} \bibinfo{volume}{624},
  \bibinfo{pages}{833--848}.
\newblock \URLprefix
  \url{https://www.sciencedirect.com/science/article/pii/S0020025522015833},
  \DOIprefix\doi{https://doi.org/10.1016/j.ins.2022.12.088}.
\bibitem[{Li et~al.(2019)Li, Jin, Xuan, Zhou, Chen, Wang and
  Yan}]{Logsparse19NIPS}
\bibinfo{author}{Li, S.}, \bibinfo{author}{Jin, X.}, \bibinfo{author}{Xuan,
  Y.}, \bibinfo{author}{Zhou, X.}, \bibinfo{author}{Chen, W.},
  \bibinfo{author}{Wang, Y.X.}, \bibinfo{author}{Yan, X.},
  \bibinfo{year}{2019}.
\newblock \bibinfo{title}{Enhancing the locality and breaking the memory
  bottleneck of {T}ransformer on time series forecasting}, in:
  \bibinfo{booktitle}{Proceedings of the Conference on Neural Information
  Processing Systems (NeurIPS)}.
\bibitem[{Lin et~al.(2020)Lin, Koprinska and Rana}]{Yang20ICONIP}
\bibinfo{author}{Lin, Y.}, \bibinfo{author}{Koprinska, I.},
  \bibinfo{author}{Rana, M.}, \bibinfo{year}{2020}.
\newblock \bibinfo{title}{Spring{N}et: Transformer and spring dtw for time
  series forecasting}, in: \bibinfo{booktitle}{Proceedings of the International
  Conference on Neural Information Processing (ICONIP)}.
\bibitem[{Lin et~al.(2021)Lin, Koprinska and Rana}]{9679135}
\bibinfo{author}{Lin, Y.}, \bibinfo{author}{Koprinska, I.},
  \bibinfo{author}{Rana, M.}, \bibinfo{year}{2021}.
\newblock \bibinfo{title}{Ssdnet: State space decomposition neural network for
  time series forecasting}, in: \bibinfo{booktitle}{2021 IEEE International
  Conference on Data Mining (ICDM)}, pp. \bibinfo{pages}{370--378}.
\newblock \DOIprefix\doi{10.1109/ICDM51629.2021.00048}.
\bibitem[{Lv et~al.(2022)Lv, Peng, Hu and Wang}]{LV2022994}
\bibinfo{author}{Lv, S.X.}, \bibinfo{author}{Peng, L.}, \bibinfo{author}{Hu,
  H.}, \bibinfo{author}{Wang, L.}, \bibinfo{year}{2022}.
\newblock \bibinfo{title}{Effective machine learning model combination based on
  selective ensemble strategy for time series forecasting}.
\newblock \bibinfo{journal}{Information Sciences} \bibinfo{volume}{612},
  \bibinfo{pages}{994--1023}.
\newblock \DOIprefix\doi{https://doi.org/10.1016/j.ins.2022.09.002}.
\bibitem[{Oreshkin et~al.(2020)Oreshkin, Carpov, Chapados and Bengio}]{N-BEATS}
\bibinfo{author}{Oreshkin, B.N.}, \bibinfo{author}{Carpov, D.},
  \bibinfo{author}{Chapados, N.}, \bibinfo{author}{Bengio, Y.},
  \bibinfo{year}{2020}.
\newblock \bibinfo{title}{N-{BEATS}: Neural basis expansion analysis for
  interpretable time series forecasting}, in: \bibinfo{booktitle}{Proceedings
  of the International Conference on Learning Representations (ICLR)}.
\bibitem[{Pham et~al.(2022)Pham, Pedrycz and Vo}]{PHAM2022117514}
\bibinfo{author}{Pham, P.}, \bibinfo{author}{Pedrycz, W.}, \bibinfo{author}{Vo,
  B.}, \bibinfo{year}{2022}.
\newblock \bibinfo{title}{Dual attention-based sequential auto-encoder for
  covid-19 outbreak forecasting: A case study in vietnam}.
\newblock \bibinfo{journal}{Expert Systems with Applications}
  \bibinfo{volume}{203}, \bibinfo{pages}{117514}.
\newblock \URLprefix
  \url{https://www.sciencedirect.com/science/article/pii/S0957417422008417},
  \DOIprefix\doi{https://doi.org/10.1016/j.eswa.2022.117514}.
\bibitem[{Qiu et~al.(2017)Qiu, Zhang, {Nagaratnam Suganthan} and
  Amaratunga}]{QIU2017249}
\bibinfo{author}{Qiu, X.}, \bibinfo{author}{Zhang, L.},
  \bibinfo{author}{{Nagaratnam Suganthan}, P.}, \bibinfo{author}{Amaratunga,
  G.A.}, \bibinfo{year}{2017}.
\newblock \bibinfo{title}{Oblique random forest ensemble via least square
  estimation for time series forecasting}.
\newblock \bibinfo{journal}{Information Sciences} \bibinfo{volume}{420},
  \bibinfo{pages}{249--262}.
\newblock \URLprefix
  \url{https://www.sciencedirect.com/science/article/pii/S0020025517309076},
  \DOIprefix\doi{https://doi.org/10.1016/j.ins.2017.08.060}.
\bibitem[{Ran et~al.(2020)Ran, Lin, Li and Zhou}]{SemiAR20ACL}
\bibinfo{author}{Ran, Q.}, \bibinfo{author}{Lin, Y.}, \bibinfo{author}{Li, P.},
  \bibinfo{author}{Zhou, J.}, \bibinfo{year}{2020}.
\newblock \bibinfo{title}{Learning to recover from multi-modality errors for
  non-autoregressive neural machine translation}, in:
  \bibinfo{booktitle}{Proceedings of the Annual Meeting of the Association for
  Computational Linguistics (ACL)}.
\bibitem[{Rangapuram et~al.(2018)Rangapuram, Seeger, Gasthaus, Stella, Wang and
  Januschowski}]{DeepSSM18NIPS}
\bibinfo{author}{Rangapuram, S.S.}, \bibinfo{author}{Seeger, M.W.},
  \bibinfo{author}{Gasthaus, J.}, \bibinfo{author}{Stella, L.},
  \bibinfo{author}{Wang, Y.}, \bibinfo{author}{Januschowski, T.},
  \bibinfo{year}{2018}.
\newblock \bibinfo{title}{Deep state space models for time series forecasting},
  in: \bibinfo{booktitle}{Proceedings of the Conference on Neural Information
  Processing Systems (NeurIPS)}.
\bibitem[{Rasul et~al.(2021)Rasul, Seward, Schuster and
  Vollgraf}]{rasul2021autoregressive}
\bibinfo{author}{Rasul, K.}, \bibinfo{author}{Seward, C.},
  \bibinfo{author}{Schuster, I.}, \bibinfo{author}{Vollgraf, R.},
  \bibinfo{year}{2021}.
\newblock \bibinfo{title}{Autoregressive denoising diffusion models for
  multivariate probabilistic time series forecasting}, in:
  \bibinfo{booktitle}{Proceedings of the International Conference on Machine
  Learning (ICML)}.
\bibitem[{Ren et~al.(2020)Ren, Liu, Tan, Zhao, Zhao and Liu}]{Ren20ACL}
\bibinfo{author}{Ren, Y.}, \bibinfo{author}{Liu, J.}, \bibinfo{author}{Tan,
  X.}, \bibinfo{author}{Zhao, Z.}, \bibinfo{author}{Zhao, S.},
  \bibinfo{author}{Liu, T.Y.}, \bibinfo{year}{2020}.
\newblock \bibinfo{title}{A study of non-autoregressive model for sequence
  generation}, in: \bibinfo{booktitle}{Proceedings of the Annual Meeting of the
  Association for Computational Linguistics (ACL)}.
\bibitem[{Salinas et~al.(2020)Salinas, Flunkert, Gasthaus and
  Januschowski}]{DeepAR20}
\bibinfo{author}{Salinas, D.}, \bibinfo{author}{Flunkert, V.},
  \bibinfo{author}{Gasthaus, J.}, \bibinfo{author}{Januschowski, T.},
  \bibinfo{year}{2020}.
\newblock \bibinfo{title}{Deep{AR}: Probabilistic forecasting with
  autoregressive recurrent networks}.
\newblock \bibinfo{journal}{International Journal of Forecasting}
  \bibinfo{volume}{36}, \bibinfo{pages}{1181 -- 1191}.
\bibitem[{Sohn et~al.(2015)Sohn, Lee and Yan}]{cVAE}
\bibinfo{author}{Sohn, K.}, \bibinfo{author}{Lee, H.}, \bibinfo{author}{Yan,
  X.}, \bibinfo{year}{2015}.
\newblock \bibinfo{title}{Learning structured output representation using deep
  conditional generative models}, in: \bibinfo{booktitle}{Proceedings of the
  Conference on Neural Information Processing Systems (NeurIPS)}.
\bibitem[{{Taieb} and {Atiya}(2016)}]{Taieb16}
\bibinfo{author}{{Taieb}, S.B.}, \bibinfo{author}{{Atiya}, A.F.},
  \bibinfo{year}{2016}.
\newblock \bibinfo{title}{A bias and variance analysis for multistep-ahead time
  series forecasting}.
\newblock \bibinfo{journal}{IEEE Transactions on Neural Networks and Learning
  Systems} \bibinfo{volume}{27}, \bibinfo{pages}{62--76}.
\bibitem[{Tong et~al.(2023)Tong, Xie, Yang, Zhang and Zhao}]{TONG2023119410}
\bibinfo{author}{Tong, J.}, \bibinfo{author}{Xie, L.}, \bibinfo{author}{Yang,
  W.}, \bibinfo{author}{Zhang, K.}, \bibinfo{author}{Zhao, J.},
  \bibinfo{year}{2023}.
\newblock \bibinfo{title}{Enhancing time series forecasting: A hierarchical
  transformer with probabilistic decomposition representation}.
\newblock \bibinfo{journal}{Information Sciences} \bibinfo{volume}{647},
  \bibinfo{pages}{119410}.
\newblock \URLprefix
  \url{https://www.sciencedirect.com/science/article/pii/S0020025523009957},
  \DOIprefix\doi{https://doi.org/10.1016/j.ins.2023.119410}.
\bibitem[{Tong et~al.()Tong, Xie and Zhang}]{doi:10.1137/1.9781611977653.ch54}
\bibinfo{author}{Tong, J.}, \bibinfo{author}{Xie, L.}, \bibinfo{author}{Zhang,
  K.}, .
\newblock \bibinfo{title}{Probabilistic decomposition transformer for time
  series forecasting}, in: \bibinfo{booktitle}{Proceedings of the 2023 SIAM
  International Conference on Data Mining (SDM)}, pp.
  \bibinfo{pages}{478--486}.
\newblock \DOIprefix\doi{10.1137/1.9781611977653.ch54}.
\bibitem[{Vaswani et~al.(2017)Vaswani, Shazeer, Parmar, Uszkoreit, Jones,
  Gomez, Kaiser and Polosukhin}]{Transformer17NIPS}
\bibinfo{author}{Vaswani, A.}, \bibinfo{author}{Shazeer, N.},
  \bibinfo{author}{Parmar, N.}, \bibinfo{author}{Uszkoreit, J.},
  \bibinfo{author}{Jones, L.}, \bibinfo{author}{Gomez, A.N.},
  \bibinfo{author}{Kaiser, {\L}.}, \bibinfo{author}{Polosukhin, I.},
  \bibinfo{year}{2017}.
\newblock \bibinfo{title}{Attention is all you need}, in:
  \bibinfo{booktitle}{Proceedings of the Conference on Neural Information
  Processing Systems (NeurIPS)}.
\bibitem[{Wang et~al.(2018)Wang, Zhang and Chen}]{SemiAR18EMNLP}
\bibinfo{author}{Wang, C.}, \bibinfo{author}{Zhang, J.}, \bibinfo{author}{Chen,
  H.}, \bibinfo{year}{2018}.
\newblock \bibinfo{title}{Semi-autoregressive neural machine translation}, in:
  \bibinfo{booktitle}{Proceedings of the Conference on Empirical Methods in
  Natural Language Processing (EMNLP)}.
\bibitem[{Wang et~al.(2022a)Wang, Pei, Zhu, Zhang, Huang, Zhai and
  Zhang}]{WANG2022262}
\bibinfo{author}{Wang, R.}, \bibinfo{author}{Pei, X.}, \bibinfo{author}{Zhu,
  J.}, \bibinfo{author}{Zhang, Z.}, \bibinfo{author}{Huang, X.},
  \bibinfo{author}{Zhai, J.}, \bibinfo{author}{Zhang, F.},
  \bibinfo{year}{2022}a.
\newblock \bibinfo{title}{Multivariable time series forecasting using model
  fusion}.
\newblock \bibinfo{journal}{Information Sciences} \bibinfo{volume}{585},
  \bibinfo{pages}{262--274}.
\newblock \URLprefix
  \url{https://www.sciencedirect.com/science/article/pii/S0020025521011452},
  \DOIprefix\doi{https://doi.org/10.1016/j.ins.2021.11.025}.
\bibitem[{Wang et~al.(2019)Wang, Smola, Maddix, Gasthaus, Foster and
  Januschowski}]{DeepFactors}
\bibinfo{author}{Wang, Y.}, \bibinfo{author}{Smola, A.},
  \bibinfo{author}{Maddix, D.}, \bibinfo{author}{Gasthaus, J.},
  \bibinfo{author}{Foster, D.}, \bibinfo{author}{Januschowski, T.},
  \bibinfo{year}{2019}.
\newblock \bibinfo{title}{Deep factors for forecasting}, in:
  \bibinfo{booktitle}{Proceedings of the International Conference on Machine
  Learning (ICML)}.
\bibitem[{Wang et~al.(2022b)Wang, Yu, Homenda, Pedrycz, Tang, Jastrzebska and
  Li}]{9763011}
\bibinfo{author}{Wang, Y.}, \bibinfo{author}{Yu, F.}, \bibinfo{author}{Homenda,
  W.}, \bibinfo{author}{Pedrycz, W.}, \bibinfo{author}{Tang, Y.},
  \bibinfo{author}{Jastrzebska, A.}, \bibinfo{author}{Li, F.},
  \bibinfo{year}{2022}b.
\newblock \bibinfo{title}{The trend-fuzzy-granulation-based adaptive fuzzy
  cognitive map for long-term time series forecasting}.
\newblock \bibinfo{journal}{IEEE Transactions on Fuzzy Systems}
  \bibinfo{volume}{30}, \bibinfo{pages}{5166--5180}.
\newblock \DOIprefix\doi{10.1109/TFUZZ.2022.3169624}.
\bibitem[{Wang et~al.(2023)Wang, Gao, Wang and Chen}]{WANG2023122504}
\bibinfo{author}{Wang, Z.}, \bibinfo{author}{Gao, R.}, \bibinfo{author}{Wang,
  P.}, \bibinfo{author}{Chen, H.}, \bibinfo{year}{2023}.
\newblock \bibinfo{title}{A new perspective on air quality index time series
  forecasting: A ternary interval decomposition ensemble learning paradigm}.
\newblock \bibinfo{journal}{Technological Forecasting and Social Change}
  \bibinfo{volume}{191}, \bibinfo{pages}{122504}.
\newblock \URLprefix
  \url{https://www.sciencedirect.com/science/article/pii/S0040162523001890},
  \DOIprefix\doi{https://doi.org/10.1016/j.techfore.2023.122504}.
\bibitem[{Wen et~al.(2018)Wen, Torkkola, Narayanaswamy and Madeka}]{MQRNN18}
\bibinfo{author}{Wen, R.}, \bibinfo{author}{Torkkola, K.},
  \bibinfo{author}{Narayanaswamy, B.}, \bibinfo{author}{Madeka, D.},
  \bibinfo{year}{2018}.
\newblock \bibinfo{title}{A multi-horizon quantile recurrent forecaster}, in:
  \bibinfo{booktitle}{Proceedings of the Conference on Neural Information
  Processing Systems (NeurIPS), Time Series Workshop}.
\bibitem[{Wu et~al.(2020)Wu, Xiao, Ding, Zhao, Wei and Huang}]{AST20NIPS}
\bibinfo{author}{Wu, S.}, \bibinfo{author}{Xiao, X.}, \bibinfo{author}{Ding,
  Q.}, \bibinfo{author}{Zhao, P.}, \bibinfo{author}{Wei, Y.},
  \bibinfo{author}{Huang, J.}, \bibinfo{year}{2020}.
\newblock \bibinfo{title}{Adversarial sparse transformer for time series
  forecasting}, in: \bibinfo{booktitle}{Proceedings of the Conference on Neural
  Information Processing Systems (NeurIPS)}.
\bibitem[{Zhou et~al.(2021a)Zhou, Zhang, Peng, Zhang, Li, Xiong and
  Zhang}]{Informer21}
\bibinfo{author}{Zhou, H.}, \bibinfo{author}{Zhang, S.}, \bibinfo{author}{Peng,
  J.}, \bibinfo{author}{Zhang, S.}, \bibinfo{author}{Li, J.},
  \bibinfo{author}{Xiong, H.}, \bibinfo{author}{Zhang, W.},
  \bibinfo{year}{2021}a.
\newblock \bibinfo{title}{Informer: Beyond efficient transformer for long
  sequence time-series forecasting}, in: \bibinfo{booktitle}{Proceedings of the
  Association for the Advancement of Artificial Intelligence (AAAI)}.
\bibitem[{Zhou et~al.(2021b)Zhou, Zhang, Hu and Wang}]{Zhou21Semi}
\bibinfo{author}{Zhou, Y.}, \bibinfo{author}{Zhang, Y.}, \bibinfo{author}{Hu,
  Z.}, \bibinfo{author}{Wang, M.}, \bibinfo{year}{2021}b.
\newblock \bibinfo{title}{Semi-autoregressive transformer for image
  captioning}.
\newblock \bibinfo{journal}{arXiv preprint arXiv:2106.09436} .

\end{thebibliography}

\end{document}